\documentclass[3p,12pt]{elsarticle}
\usepackage{hyperref}
\usepackage{textcomp}
\usepackage[export]{adjustbox}
\usepackage{amssymb}
\usepackage{pifont}
\newcommand{\cmark}{\ding{51}}%
\newcommand{\xmark}{\ding{55}}%
\usepackage[normalem]{ulem}

\usepackage{times}
\usepackage{xcolor}
\usepackage{soul}
\usepackage{makecell}
\usepackage{graphicx}
\usepackage{amsmath}
\usepackage{amsthm}
\usepackage{amssymb}
\usepackage{booktabs}
\usepackage{algorithm}
\usepackage{algorithmic}
\usepackage[small]{caption}
\usepackage{amssymb}
\usepackage{multirow}
\usepackage{longtable}
\usepackage{tabularx,booktabs}
\usepackage{pdfpages}
\usepackage{forest}
\usepackage{comment}

\usepackage{tikz}

\newcommand{\dung}[1]{{\color{black}{#1}}}
\newcommand{\tuan}[1]{{\color{black}{#1}}}

\usetikzlibrary{positioning, shapes.geometric}

\usepackage[figuresright]{rotating}

\begin{document}

\begin{frontmatter}




\title{Backdoor Attacks and Defenses in Federated Learning: \tuan{Survey}, Challenges and Future Research Directions}

\author[label1,label2]{Thuy Dung Nguyen}
 \ead {dung.nt2@vinuni.edu.vn}
\author[label1,label2]{Nguyen Tuan}
 \ead {tuan.nm@vinuni.edu.vn}
\author[label3]{Phi Le Nguyen}
 \ead {lenp@soict.hust.edu.vn}
\author[label1,label2]{Hieu H. Pham}
 \ead {hieu.ph@vinuni.edu.vn}
\author[label1]{Khoa Doan}
 \ead {khoa.dd@vinuni.edu.vn}
\author[label1]{Kok-Seng Wong\corref{cor1}}
\cortext[cor1]{Corresponding authors}
 \ead {wong.ks@vinuni.edu.vn}

\address[label1]{College of Engineering \& Computer Science, VinUniversity, Hanoi, Vietnam}
\address[label2]{VinUni-Illinois Smart Health Center, VinUniversity, Hanoi, Vietnam}
\address[label3]{Hanoi University of Science and Technology, Hanoi, Vietnam}



\begin{abstract}
Federated learning (FL) is a machine learning (ML) approach that allows the use of distributed data without compromising personal privacy. However, the heterogeneous distribution of data among clients in FL can make it difficult for the orchestration server to validate the integrity of local model updates, making FL vulnerable to various threats, including backdoor attacks. Backdoor attacks involve the insertion of malicious functionality into a \tuan{targeted} model through poisoned updates from malicious clients. These attacks can cause the global model to misbehave on specific inputs while appearing normal in other cases. Backdoor attacks have received significant attention in the literature due to their potential to impact real-world deep learning \tuan{applications}. However, they have not been thoroughly studied in the context of FL. In this survey, we provide a comprehensive \tuan{survey} of current backdoor attack strategies and defenses in FL, including a \tuan{comprehensive analysis} of different approaches. 
We also discuss the challenges and potential future directions for attacks and defenses in the context of FL.

\end{abstract}


\end{frontmatter}


\section{Introduction}
\label{sec:intro}
Artificial intelligence (AI) and machine learning (ML) can analyze large amounts of data, identify patterns, make decisions, improve efficiency, and solve complex problems in various fields. These technologies have the potential to greatly improve industries such as healthcare, finance, and education~\cite{ai_ml_opportunities_2018}. The success of many deployed ML systems crucially hinges on the availability of high-quality data. However, a single entity does not own all the data it needs to train the ML model. Specifically, the valuable data examples or features are scattered in different organizations or entities. For example, medical images sit in data silos, and privacy concerns limit data sharing for ML tasks. Consequently, large amounts and diverse medical images from different hospitals are not fully exploited by ML.
Federated learning (FL)~\citep{McMahan2016FederatedLO, Bonawitz2019TowardsFL} which was introduced by Google is a decentralized ML \tuan{paradigm} that allows multiple devices to train a global model collaboratively without compromising data privacy by storing data locally on end-user devices. The orchestration server collects and aggregates model updates from the participating clients to calculate a global model update \tuan{which will be sent to the clients} in the next training round. Due to its advantages, FL has been widely used in \tuan{various fields including} \tuan{computer vision (CV)~\cite{Doshi2022FederatedLD,Becking2022AdaptiveDF}}, natural language processing (NLP)~\citep{Chen2019FederatedLO,Lin2021FedNLPAR}, healthcare~\citep{Xu2021FederatedLF,Prayitno2021ASR,Gupta2021HierarchicalFL,Liu2021FederatedLA}, and Internet of Things (IoT)~\citep{Liu2021DeepAD,Kholod2021OpenSourceFL,Nguyen2021EfficientFL}. 
However, the decentralized \tuan{nature} of FL makes it more challenging to verify the trustworthiness of each participant, 
\tuan{leading to a vulnerability to various attacks}
~\cite{Lyu2020ThreatsTF}.

\begin{figure*}[tb]
    \centering
    \begin{minipage}{0.325\linewidth}
        \includegraphics[width=1.0\textwidth]{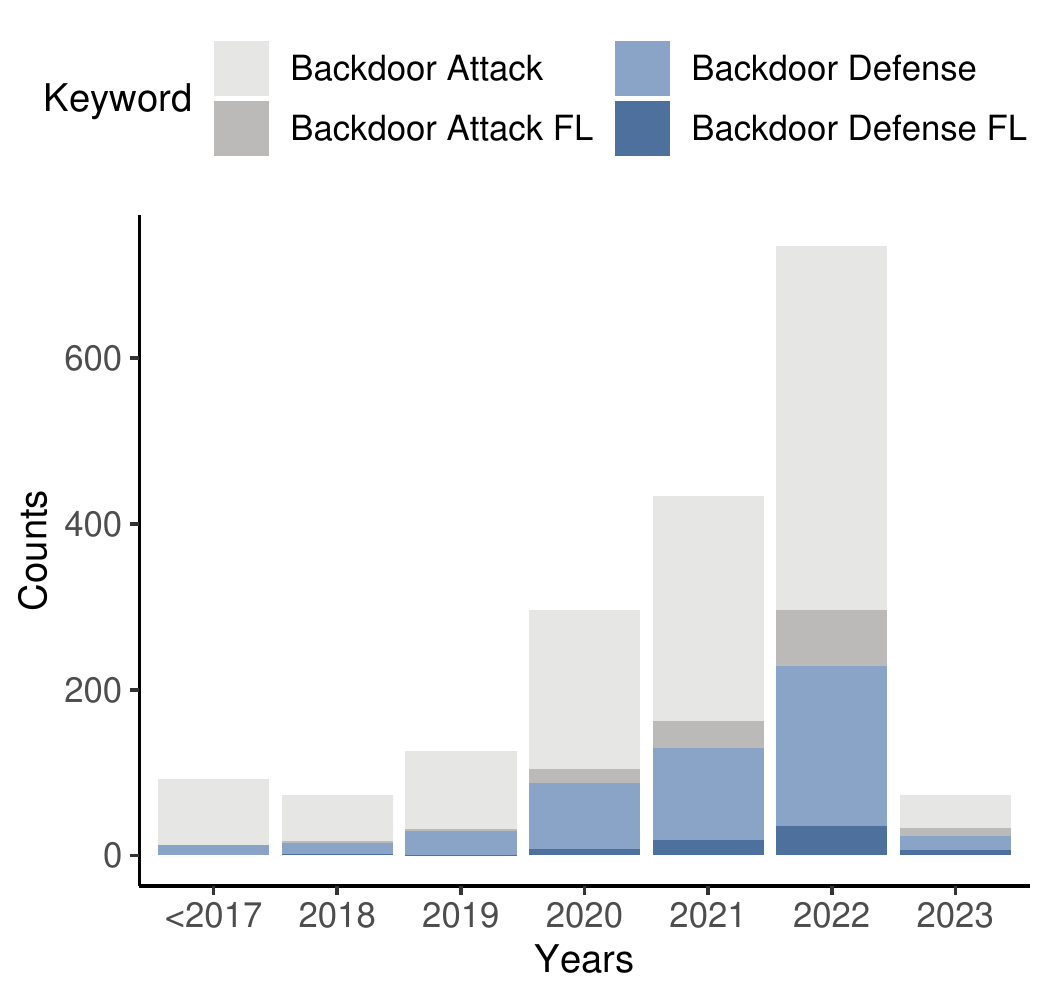}
        \caption{Frequency of backdoor-related keywords appeared on titles or abstracts of publications by \textit{app.dimensions.ai}. \textit{Note}: The data was collected from 2014 to 1 February 2023.}
        \label{fig:count_works}
    \end{minipage}
\hfill
\begin{minipage}{0.63\linewidth}
    \centering
	\includegraphics[width=1.0\textwidth]{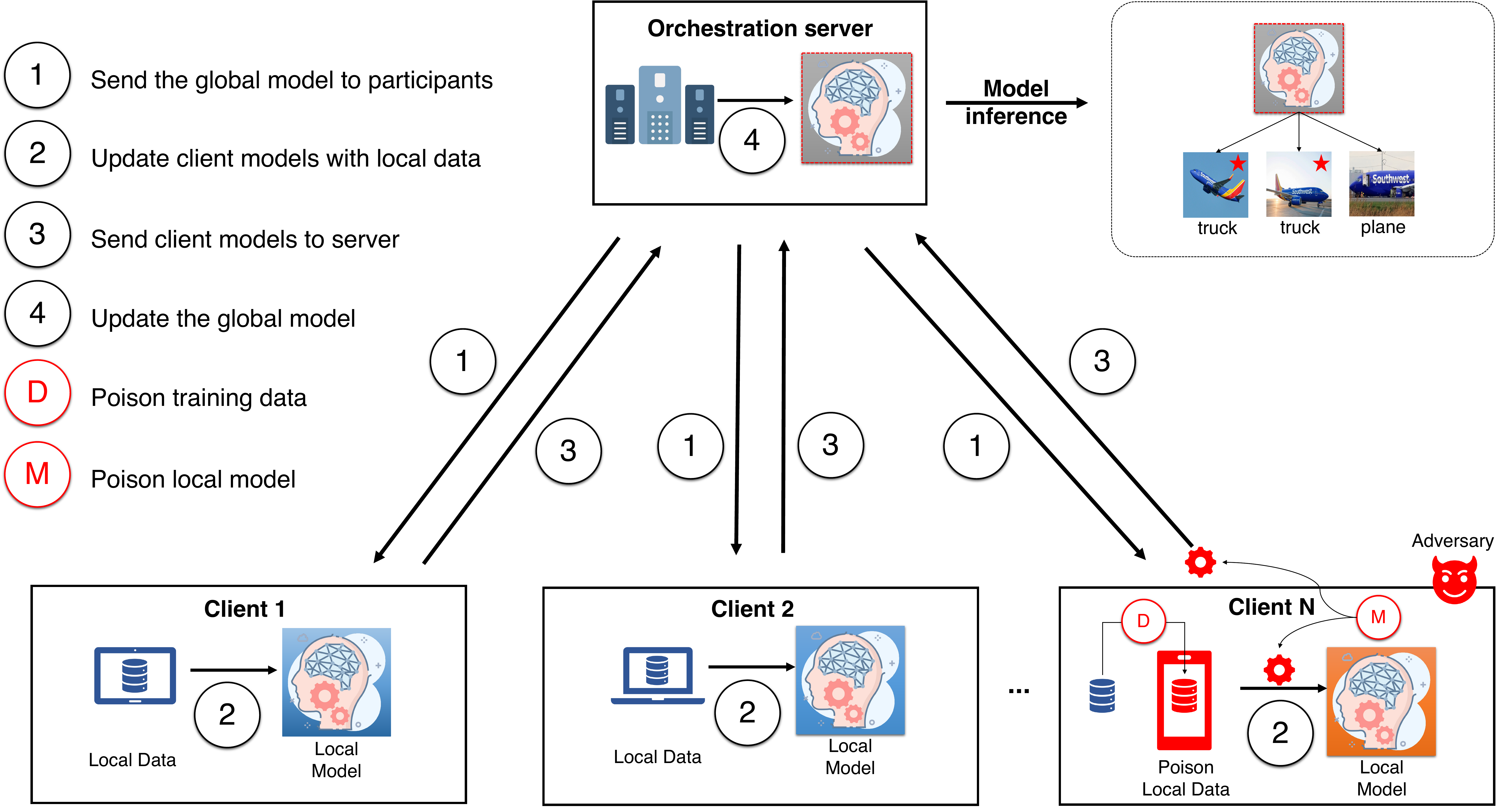}
\caption{Overview of Backdoor Attacks in Centralized FL. There are malicious participants who attempt to submit poisoned updates to backdoor the model. At the inference stage, the global model misbehaves on triggered-inputs, i.e., plane images with red star.}
	\label{fig:overview}
\end{minipage}
\hfill
\vspace{-0.3cm}
\end{figure*}

Among the attacks operating against FL, backdoor attacks are raising concerns due to the possibility of stealthily injecting a malevolent behavior within the global model~\cite{bagdasaryan2020backdoor,wang2020attack}. In particular, a trigger in test-time input forces the backdoored model to behave in a specific manner that the attacker desires while ensuring that the poisoned model behaves normally without triggers. \dung{As shown in Figure~\ref{fig:count_works}, the number of works focused on the backdoor attack fields is increasing exponentially in the literature, indicating the importance of this topic for the security of ML and FL.} To implant the backdoor, most existing backdoor attacks target centralized FL, in which the orchestration server is assumed to be honest, and there are several malicious participants (as illustrated in Figure~\ref{fig:overview}).
Unlike backdoor attack in ML, an adversary in FL can insert poisons at various stages of the training pipeline (i.e., poisoning data and poisoning model), and attacks are not constrained to be ``clean-label", making it more challenging to design a backdoor-robust FL scheme. Indeed, much effort was devoted to demonstrating that FL is vulnerable to the backdoor attack, and with a carefully designed attack scheme, the adversary can successfully manipulate the global model without being detected~\citep{bagdasaryan2020backdoor,wang2020attack,xie2020dba,Neurotoxin}. The impacts of backdoor attacks can be seen \tuan{in many FL scenarios across research fields such as CV~\citep{Wenger2021PhysicalBA, HTBD2018}, NLP\citep{wang2020attack,Sun2019CanYR}, and IoT networks~\citep{Nguyen2020PoisoningIoT}. In addition, these attacks can also affect application domains such as healthcare systems~\citep{jin2022backdoor}.}
\tuan{For instance, in healthcare, FL is used to train ML models for various applications, such as predicting patient outcomes using medical records. However, in the case of a backdoor attack, the ML model could potentially make incorrect predictions, as was demonstrated in a backdoor attack on a deep learning model used for skin lesion classification, which could have serious consequences for patients' health~\cite{9880076}}.


\tuan{To cope with} new threats posed by backdoor attacks, many FL defenses have been proposed\tuan{~\citep{fung2018mitigating, shen2016auror, Sun2019CanYR, wu2022federated, Wu2020MitigatingBA}}. As a result, defense mechanisms against backdoor attacks in FL can be conducted in different phases of the learning process, including pre-aggregation, in-aggregation, and post-aggregation. \tuan{Defenses in the pre-aggregation process}~\cite{fung2018mitigating, shen2016auror, Nguyen2022-FLAME, rieger2022deepsight} aim to identify and remove (or reduce) the impact of malicious updates before the global update phase happens. In-aggregation \tuan{defense} techniques~\citep{Sun2019CanYR, Pillutla2022RobustAF, blanchard2017machine, ozdayi2021defending} use more robust aggregation operators to alleviate the \tuan{backdoor effects} while global model updating procedure is conducted. Meanwhile, the post-aggregation defense techniques~\citep{wu2022federated, Wu2020MitigatingBA} aim to repair backdoored models after \tuan{completing the FL training process.}
However, existing countermeasures are mainly attack-driven, i.e., they can only defend against well-known attack techniques, and an adversary who is aware of the existence of these defenses can circumvent them~\cite{Bhagoji2019AnalyzingFL, wang2020attack}. One explanation for this is that backdoor defenses are developed mostly \tuan{based} on observations and assumptions, rather than a thorough understanding of attack methodologies and learning algorithms. As a result, a comprehensive and in-depth survey is required to better understand \tuan{the backdoor attacks and defenses in FL.}
\renewcommand{\arraystretch}{1.25}

\begin{table}[tbh]
\centering
\caption{{A Summary of Existing Surveys Related to FL Backdoor Attacks}}
\vspace{1em}
\label{tab:survey-summary}
\resizebox{0.95\textwidth}{!}{
\begin{tabular}{@{}p{0.3\textwidth}cccccccccc@{}}
\toprule
         &  
         & \multicolumn{4}{c}{Main focus}                                   
                & \multicolumn{5}{c}{Survey dimensions}                                
\\ 
\cmidrule(l){3-6} \cmidrule(l){7-11}
\multicolumn{1}{c}{Survey paper} & Year 
        & \multicolumn{1}{c}{\begin{tabular}[c]{@{}c@{}}Privacy/security \\ threats\end{tabular}} 
        & \multicolumn{1}{c}{\begin{tabular}[c]{@{}c@{}}Poisoning\\ attacks\end{tabular}}
        & \begin{tabular}[c]{@{}c@{}}Backdoor \\ attacks\end{tabular} 
        & \begin{tabular}[c]{@{}c@{}}Backdoor \\ defenses\end{tabular} 
        
    & \multicolumn{1}{c}{\begin{tabular}[c]{@{}c@{}}Different \\FL category\end{tabular}}
                & \multicolumn{1}{c}{\begin{tabular}[c]{@{}c@{}}Technique\\ driven\end{tabular}} 
                & \multicolumn{1}{c}{\begin{tabular}[c]{@{}c@{}}Inter\\connection\end{tabular}} 
                & \multicolumn{1}{c}{\begin{tabular}[c]{@{}c@{}}Evaluation \\ metrics\end{tabular}} 
                & \begin{tabular}[c]{@{}c@{}} Backdoor\\ applicability\end{tabular} \\ \midrule
                
Data poisoning attacks\cite{Goldblum2022-wq} & 2022 &
& \checkmark & & &
& \checkmark & & & \checkmark\\
Backdoor attacks and defenses\cite{Li2022BackdoorLA} & 2022 & & & \checkmark & \checkmark & & \checkmark & \checkmark\\
 Backdoor attacks and defenses in FL\cite{ba-fl-survey-2022} & 2022 & &  & \checkmark & & & \checkmark &\\
Poisoning attacks and countermeasures\cite{Tian2022ACS} & 2022 & & \checkmark & & &  &\checkmark & \checkmark & \checkmark & \checkmark \\
FL challenges, contributions, and trends\cite{Abdulrahman2021ASO} & 2021  & \checkmark & & & & \checkmark &\checkmark\\
 Privacy-preserving FL\cite{Yin2021ACS} & 2021 & \checkmark & & & & \checkmark & \checkmark\\
 Security and Privacy in FL\cite{Mothukuri2021ASO} & 2021 & \checkmark & & & &\checkmark & \checkmark \\
 FL security and privacy threats\cite{Gosselin2022PrivacyAS} & 2022 & \checkmark & & & & \checkmark & \checkmark \\
 Threats and attacks in FL\cite{Lyu2020ThreatsTF} & 2020 & \checkmark & & &  &  \checkmark & \checkmark &  & \\                                               
&                                                                   \\ 
 
  Backdoor poisoning attacks\cite{Chen2017TargetedBA} & 2017 & & &  \checkmark & & &\checkmark & & & \checkmark\\
  
\cmidrule(l){1-11}
\multicolumn{2}{c}{\textbf{Ours}} & & & \checkmark & \checkmark&  \checkmark  &  \checkmark & \checkmark & \checkmark & \checkmark\\
\bottomrule
\end{tabular}
}
\end{table}

\subsection{Related Surveys}
In this work, we review recent survey papers in the literature (from 2020 to 2022) by searching for relevant papers using keywords related to ``backdoor attack" and ``federated learning" in various academic databases such as IEEE Xplore, ACM Digital Library, and arXiv. We also \tuan{include} a paper from 2017~\cite{Chen2017TargetedBA}, as it was one of the first papers introducing the concept of backdoor attacks in ML. As summarized in Table~\ref{tab:survey-summary}, most existing surveys on FL are focused on privacy and security threats, and \tuan{the} backdoor attack is only considered as a specific instance of the targeted poisoning attacks~\cite{Abdulrahman2021ASO, Yin2021ACS, Mothukuri2021ASO, Gosselin2022PrivacyAS} or as a special example of robustness threats~\cite{Lyu2020ThreatsTF}. Consequently, these surveys contribute less to improving the understanding of the working mechanism of backdoor attacks and their vulnerabilities in FL. Other surveys~\cite{Goldblum2022-wq, Li2022BackdoorLA, Chen2017TargetedBA} study FL backdoor attacks as a special case of those in deep learning. However, the criteria to systematize backdoor attacks is too immense for studying FL backdoor attacks, \tuan{since} the attack methodology in FL is significantly different from attacks in centralized learning. These surveys examine FL backdoor attacks from a technique-driven perspective by reviewing state-of-the-art FL backdoor attacks and countermeasures based on their key methods and contributions. Still, they do not fully study them under the unique dimensions of FL, such as data partition strategy and participant contribution. In ~\cite{ba-fl-survey-2022}, the authors \tuan{focus on investigating} FL backdoor attacks and cutting-edge defenses. In this study, backdoor attacks are classified into data poisoning and model poisoning attacks. \tuan{In addition, the authors} review significant works corresponding to each approach and compare them in terms of their attack settings. 
\tuan{Their survey, however, falls short of assessing or demonstrating the connection between these attacks, as well as the connection between backdoor attacks and defenses.}
In addition, we also observe that the evaluation metrics and applicability of backdoor attacks in the physical world have not been discussed in the existing survey papers.

\subsection{Our Contributions}
In comparison with the previous surveys, we focus on the functioning mechanism and evolution of FL backdoor attacks from multi-perspectives: techniques, relationships, evaluation metrics, and applicability. Furthermore, we also study their efficiency and limitations under many dimensions, including adversary assumption, stealthiness, and durability, which are not included in previous works.
We also evaluate the effectiveness of defense mechanisms in terms of their robustness against various attack schemes, including physical attacks.
The main objectives of this survey are to improve the understanding of FL backdoor attacks (and their consequences) and to assist academia and industry in developing more robust FL systems. To achieve this, a new taxonomy of FL backdoor attacks and defenses is provided, as well as a discussion of future research directions from a multi-perspective viewpoint. Furthermore, a comprehensive review of the current state of the art in FL backdoor attacks and defenses is presented. The main contributions of this work are summarized as follows:
\begin{enumerate}

\item We separate FL backdoor attacks into two main categories based on the training stages in which they \tuan{happen}. The category is further divided into 13 subcategories regarding adversarial objectives and \tuan{methodologies}. Based on this, we provide a comprehensive analysis that covers a critical review and comparison of each backdoor attack strategy.

\item We review the state-of-the-art defense strategies and categorize them based on their common objectives and methodologies. In addition, we provide a comprehensive analysis of their efficiency against existing backdoor attacks and their applicability.

\item We discuss the challenges for both backdoor attacks and defenses in FL, followed by possible future works in different aspects, and demonstrate significant missing gaps that need to be addressed.

\item To the best of our knowledge, this is the first survey to assess and analyze backdoor attacks and defenses utilizing FL-specific criteria and perspectives. Our survey aims to enhance the development of more sophisticated methods and increase the understanding of backdoor threats and countermeasures, thus contributing to the building of more secure FL systems.

\end{enumerate}


The rest of the paper is organized as follows. In section \ref{sec:background}, we provide the overview of FL, backdoor attacks, and evaluation metrics, followed by attack techniques in Section \ref{sec:backdoor-techniques}. In Section \ref{sec:backdoor_defense_methodologies}, we review the defense strategies against backdoor attacks. We discuss the challenges and future directions in Section \ref{sec:backdoor_future_directions}, \tuan{and summarize the key findings} and conclusion in Section \ref{sec:conclusion}.


\section{Background}
\label{sec:background}
\subsection{Definition of Technical Terms}
This section presents concise definitions and descriptions of technical terms used in FL systems, backdoor attacks, and defenses in Table~\ref{tab:terminology}. These definitions will be consistently referred to throughout the remainder of the survey.

\begin{table*}[hbt!]
\scriptsize
\centering
\caption{Terminology Definitions }
\vspace{1em}
\label{tab:terminology}
\resizebox{0.95\textwidth}{!}{
\renewcommand{\arraystretch}{1.5}
\begin{tabular}{@{}l p{0.5\textwidth}p{0.15\textwidth}@{}}
\toprule 
\multicolumn{1}{c}{Terminology} & \multicolumn{1}{c}{Definition}                       & Exchangeable Terms                      
\\ \midrule
Orchestration server & The server has the power to manage the communication and information of participating clients in the FL system & Central server, Federated server, FL server, Aggregator \\
Benign clients & Clients training with benign settings and are not controlled by any adversary & Honest clients \\
Malicious clients & Clients training with poisoning settings and are controlled by an adversary & Compromised clients, Dishonest clients\\
Poisoned sample & The modified training sample used in poisoning-based backdoor attacks was used to implant backdoor(s) in the model during the training phase & N/A\\
Trigger & The pattern is embedded in the poisoned samples and it is used to activate the hidden backdoor(s) & Backdoor key \\
Backdoor target & The objective of the backdoor attack which describes the specific characteristics of poisoned samples and the corresponding targeted class or label & Adversarial task, Backdoor task \\
Black-box attack & The adversary has no knowledge about the target model, and is only able to replace their local data set & N/A \\
White-box attack & The adversary is able to manipulate the training data and local model training's parameters & N/A \\
Full-box attack & The adversary has complete control over the local training process and can replace the training protocol, i.e., using sub-training process to learn the transformation model which outputs backdoored samples & N/A \\
Continuous attack & The backdoor attacks are carried out continuously during the training process, either by all communication rounds or a portion of them & N/A \\
Single-shot attack & During the training process, the malicious client(s) are selected in only a single round of training & N/A \\
Collusion & The adversary controls more than one clients and requires their poisoned updates to facilitate the backdoor attack & N/A \\
Poisoned Model Rate & The ratio of malicious clients per total in FL & PMR \\
\bottomrule
\end{tabular}}
\end{table*}

\subsection{Overview of Federated Learning}
FL has recently received considerable attention and is becoming a popular ML framework that allows clients to train machine learning models in a decentralized fashion without sharing any private dataset. In the FL framework, data for learning tasks are acquired and processed locally at the edge node, and only the updated ML parameters are transmitted to the central orchestration server for aggregation. In general, FL involves the following main steps (as illustrated in Steps 1 to 4 in Figure~\ref{fig:overview}): 
\begin{itemize}
    \item \textit{Step 1 (FL Initialization}): the central orchestration server $\mathcal{S}$ will first initiate the weight of the global model and the hyperparameters such as the number of FL rounds and local epochs, size of the selected clients for each round, and the local learning rate.  
    \item \textit{Step 2 (Local Model Training}): all selected clients $\mathcal{C}_1,\mathcal{C}_2,...,\mathcal{C}_m$\tuan{, where $\mathcal{C}_i$ represents client number $i$, receive the current global weight from $\mathcal{S}$}. Next, each $\mathcal{C}_i$ updates its local model parameters $\textbf{w}^t_i$ using its local dataset, $\mathcal{D}_i$, where $t$ denotes the current iteration round. 
    \item \textit{Step 3 (Local Model Update}): Upon the completion of the local training, all selected clients send the local weight to $\mathcal{S}$ for model aggregation. 
    \item \textit{Step 4 (Global Model Aggregation and Update Phase)}: $\mathcal{S}$ aggregates the received local weights and sends back the aggregation result to the clients for the next round of training.
\end{itemize}

\dung{The aggregation techniques can produce a robust training model in some instances if we make certain assumptions about the type of attack and limit the number of malicious clients. Above all, FedAvg~\citep{mcmahan2017communication} is widely used in FL for both attack and defense scenarios, in particular in work about backdoor attacks and defenses \cite{bagdasaryan2020backdoor, Nguyen2022-FLAME, shen2016auror, nguyen2020poisoning, munoz2019byzantine, fung2020limitations}. In FedAvg, the aggregated model $\textbf{W}^{t+1}$ at round $t + 1$ is determined by taking the average of all model updates and adding them to the previous global model $\textbf{W}^{t}$ at round $t$. Despite the fact that this algorithm also allows weighting the contributions of different clients, e.g., to increase the impact of clients with a large training dataset, this also makes the system more vulnerable to manipulation, as compromised clients could exploit this to increase their impact, e.g., by lying about the size of their datasets.}
Besides FedAvg, different aggregation rules have been proposed in the literature (e.g., Krum~\cite{blanchard2017machine}, Trimmed-Mean~\cite{pmlr-v80-yin18a}, and SimFL~\cite{li2020practical}) to improve the FL performance and convergence time.

In FL settings, an attacker may attempt to compromise the integrity of the models and data used during the process of updating client models with local data, as illustrated in Figure~\ref{fig:overview}. One tactic that an attacker may employ is the model modification, in which the attacker alters the parameters of a local model on a participating client before it being sent back to the central server. Through this manipulation, the attacker can insert a ``backdoor" into the model, allowing it to produce a desired output when a specific input, also known as a trigger, is provided. Another technique that an attacker may utilize is data poisoning, in which the attacker manipulates the data to train a local model on a participating client. This can include adding specific images or patterns to the data, which can cause the model to recognize them as triggers for malicious behavior.
\begin{figure}[hbt!]
	\centering
	\includegraphics[width=\textwidth]{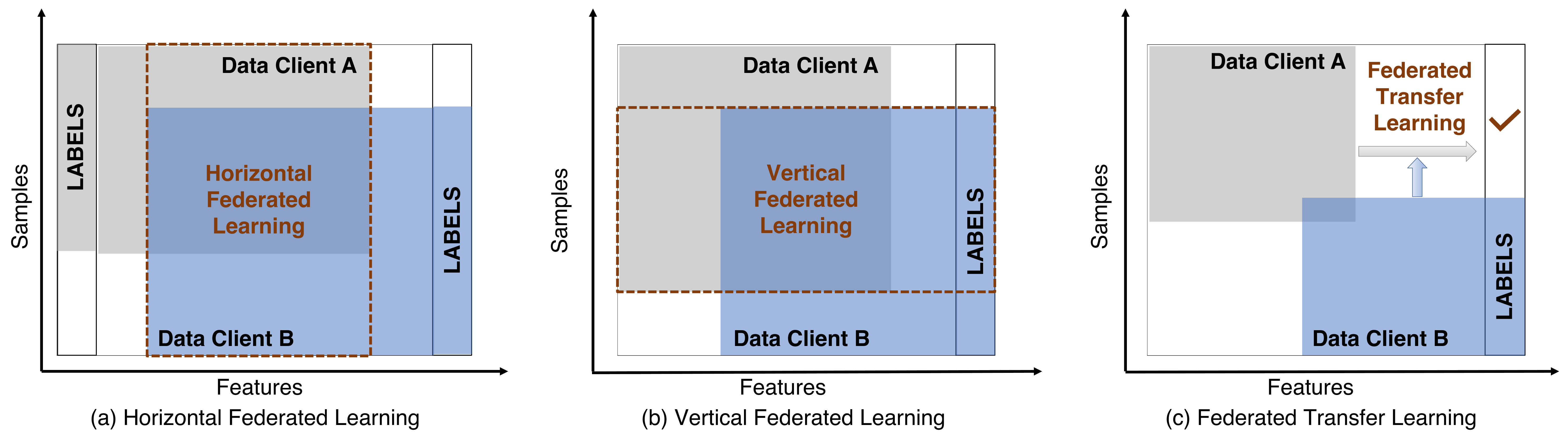} 
	\caption{Categorization of FL based on the distribution of data.}
	\label{fig:tao-fl}
\end{figure}

Based on the distribution of data features and samples among clients, FL can be categorized into horizontal federated learning (HFL)~\cite{smith2017federated}, vertical federated learning (VFL)~\cite{wu2020privacy, yang2019parallel}, and federated transfer  learning (FTL)~\cite{chen2020fedhealth}. HFL is used when different datasets share the same feature space but differ in sample IDs whereas VFL shares the same sample IDs but differs in feature space. FTL is used when different datasets do not share the same sample IDs or feature space and involve transferring knowledge from a source domain to a target domain to improve learning outcomes.
We show the overview of the FL categorization in Figure~\ref{fig:tao-fl}.

\subsection{Backdoor Attack in Federated Learning}
\dung{Backdoor attacks in FL have been studied as a potential security threat in FL systems. The main idea behind a backdoor attack in FL is to manipulate the local models in a FL setup to compromise the global model.
In these attacks, an attacker tries to introduce a trigger in one or more of the local models, such that the global model will have a specific behavior under the presence of the trigger on the inputs. In the context of autonomous driving, for instance, an attacker may desire to offer the user a backdoored street sign detector that has high accuracy for detecting street signs under normal conditions but identifies stop signs with a certain sticker as speed limit signs (e.g., a smiley face)~\cite{gu2019badnets}. 
}

\dung{A backdoor attack in FL could be formulated as a multi-objective optimization problem, where the attacker is trying to optimize the following objectives simultaneously
\begin{equation}
    \theta^* = \min_{\theta} \sum_{i \in |\mathcal{D}|}^{}\mathcal{L}(x_i, y_i)+ \sum_{i \in |\mathcal{D}_p|}^{}\mathcal{L}(\varphi(x_i), \tau(y_i)),
\end{equation}
}
\dung{in which $\mathcal{D}$ is the benign testing set representing for the main task to learn, and $\mathcal{D}_p$ is the poisoning set including the backdoored samples. These samples are manipulated by a transform function $\varphi$, which can be a non-transform function~\cite{wang2020attack} or a perturbation function~\cite{xie2020dba, xie2021crfl}. Technically, the adversary objective is to manipulate the model such that it makes distorted outputs for these poisoned sample (i.e., the model outputs $\tau(y_i)$ given $\varphi(x_i)$). \tuan{The function $\mathcal{L}$ in the expression $\mathcal{L}(\varphi(x_i), \tau(y_i))$ represents a loss function that measures the discrepancy between the predicted output $\varphi(x_i)$ and the true output $\tau(y_i)$ for a given input sample $(x_i, y_i)$}
At the same time, to ensure the stealthiness, the performance of the model on non-backdoored samples remains unchanged. In particular, model should $\theta^*$ gives true outputs for samples $x_i$ not belonging to $\mathcal{D}_p$ set.}

In contrast to backdoor attacks in centralized learning, existing backdoor attacks in FL are based on the scenario that adversaries cannot directly influence the federated model, and they poison the model by updating the backdoored updates from their compromised participants. As a result, the aggregation of updates from multiple clients may reduces the effect of an individual malicious update~\cite{bagdasaryan2020backdoor}.

\subsection{Evaluation Metrics}
The objective of an adversary's backdoor attack is to mislead the global model to produce incorrect outputs on backdoored inputs (e.g., the global model classifies images of \tuan{``green cars" as ``frogs"} in an image classification task). Therefore, the metrics used to evaluate the effectiveness of a backdoor attack are related to the attack's objective. One metric, called attack success rate (ASR)~\citep{xie2020dba}, measures the probability that the output of the backdoored model on targeted inputs matches the adversary's preference. Other term such as backdoor task accuracy~\citep{wang2020attack} refers to the same concept as ASR. In general, a higher backdoor accuracy corresponds to a higher attack success rate.

Mathematically, let $\tilde{\mathbf{\mathcal{D}}}$ be the targeted samples (e.g., images inserted trigger pattern), and the $\tau$ be the targeted class of the adversary. Since the backdoored model $\mathbf{f}_{\textbf{W}}$ is expected to misclassify $\tilde{\mathbf{\mathcal{D}}}$ as $\tau$, the ASR is calculated by
\begin{equation}
    \text{ASR} = \sum_{x\in \tilde{\mathbf{\mathcal{D}}}} \frac{ \mathbf{f}_{\mathbf{W}}(x) = \tau}{| \tilde{\mathbf{\mathcal{D}}} |}
\end{equation}

\begin{figure*}[tb]
    \centering
\begin{minipage}{0.7\textwidth}
	\centering\includegraphics[width=1.0\textwidth]{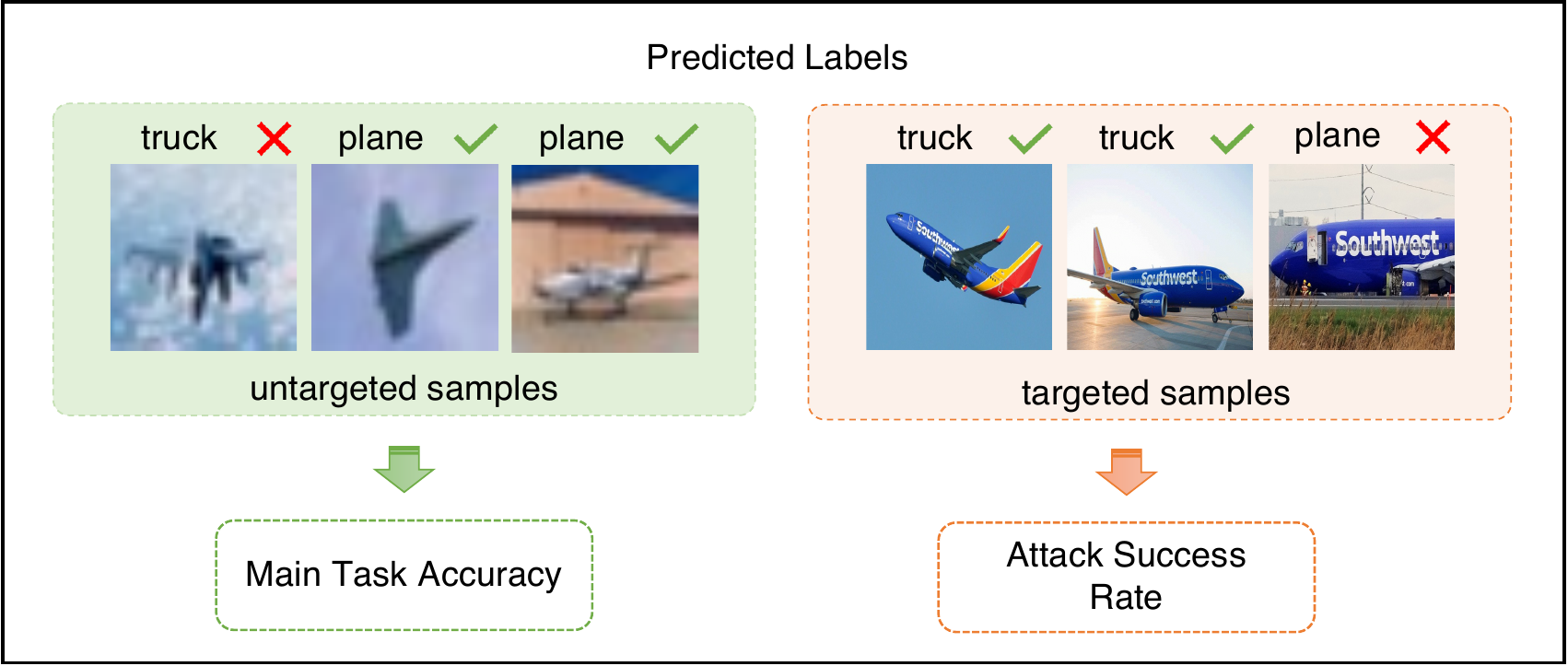} 
	\caption{Common metrics for \tuan{Backdoor Attacks and Defenses.}
	}
	\label{fig:backdoor-metrics}
\end{minipage}
\hfill
\end{figure*}

Additionally, the trained model $\mathbf{f}_{\textbf{W}}$ should produce normal outputs on benign samples (e.g., images without triggers). The model's accuracy on these samples can be measured using the metric called main task accuracy (MTA)~\citep{xie2020dba} on benign samples. This is calculated as
\begin{equation}
    \text{MTA} = \sum_{x_i\in \mathbf{\mathcal{D}}} \frac{ \mathbf{f}_{\mathbf{W}}(x_i) = y_i}{| \mathbf{\mathcal{D}} |},
\end{equation}
where $\mathbf{\mathcal{D}} := \left[ x_1^{y_1}, x_2^{y_2}, \dots,  x_{|\mathbf{\mathcal{D}}|}^{y_{|\mathbf{\mathcal{D}}|}}\right]$  is the validation set held by the aggregator, and $y_i$ is the corresponding label for sample $x_i$. In most backdoor attack strategies, the adversary is successful in planting the backdoor only if the trained model has both high MTA and significant ASR~\citep{bagdasaryan2020backdoor,wang2020attack}. A simple illustration of these two common metrics is shown in Figure~\ref{fig:backdoor-metrics}.

To evaluate the effectiveness of FL defenses against backdoor attacks, ASR and MTA which are mentioned above are widely used. In most existing defenses, the authors aimed at minimizing the ASR while not degrading the MTA. In addition, in the anomaly detection-based defenses, other metrics are employed to evaluate the accuracy in detecting malicious updates~\cite{nguyen2022flame}. In particular, they measure true positive rate (TPR) and true negative rate (TNR), which are defined as follows.
\begin{itemize}
    \item \textbf{TPR: }measures how well the defense identifies poisoned models, e.g., the ratio of the number of models correctly classified as poisoned (True Positives - TP) to the total number of models being classified as poisoned: $TPR = \frac{TP}{TP+FP}$, where FP is False Positives indicating the number of benign clients that are wrongly classified as malicious.
    \item \textbf{TNR:} measures the ratio of the number of models correctly classified as benign (True Negatives - TN) to the total number of benign models: $TNR = \frac{TN}{TN+FN}$, where FN is False Negatives indicating the number of malicious clients that are wrongly classified as benign.
\end{itemize}

\renewcommand{\arraystretch}{1.5}
\begin{table*}[hbt]
\caption{{Comparison of State-of-the-art Backdoor Attack Strategies in FL}}
\label{tab:backdoor-attack-comparision}
\resizebox{\textwidth}{!}{
\begin{tabular}{llcccccccccc}
\hline
&  & 
\multicolumn{3}{c}{Backdoor Characteristics}     & \multicolumn{3}{c}{Adversary Assumption}          & \multicolumn{3}{c}{Attack Efficiency}         \\

\cmidrule(lr){3-5} 
\cmidrule(lr){6-8} 
\cmidrule(lr){9-12}
    \multirow{1}{*}{Name}  &   
    \multirow{1}{*}{Year}  & 
    \makecell{Data \\Poisoning}  & 
    \makecell{Model \\Poisoning} & 
    \makecell{Accessibility} & \makecell{Collusion \\Required} & \makecell{Continuous \\Attack} & 
    \makecell{Converging Stage\\Constraint} & \makecell{Extended \\Durability} &
    \makecell{Stealthiness \\Consideration} &
    \multirow{1}{*}{\makecell{FL Type}} &
    \multirow{1}{*}{\makecell{Applications}} \\ 
    \hline
RE+GE~\cite{rare-Yoo2022BackdoorAI} & 2022 & \makecell{D1} & M4 & White-box & \cmark & \cmark & \xmark & \xmark & \xmark & HFL & NLP\\
CBA~\cite{Gong2022CoordinatedBA} & 2022 & D2 & -- & Full-box & \cmark & \cmark & \xmark & \xmark & \xmark & HFL & IC                                                                \\
Neurotoxin~\cite{Neurotoxin} & 2022 & -- & M4 & White-box & \xmark & \cmark & \xmark & \cmark & \xmark & HFL & \makecell[c]{NLP, IC}                                                                 \\
GRA-HE~\cite{Zou2022VFL} & 2022 & D2 & M3 & Full-box & \xmark & \cmark & \xmark & \xmark & \xmark & VFL & IC                                                                          \\
DeepMP~\cite{Zhou2021DeepMP} & 2021 & -- & M4 & White-box & \xmark & \xmark & \xmark & \cmark & \cmark & HFL & IC                                                           \\
PoisonGAN~\cite{PoisonGAN} & 2021 & D1 & -- & Full-box & \cmark & \cmark & \xmark & \xmark & \xmark & HFL & IC                                                                   \\
DBA~\cite{xie2020dba} & 2020 & \makecell{D2} & -- & White-box & \cmark & \xmark & \cmark & \xmark & \cmark & HFL & IC                                                                           \\
PFLIoT~\cite{Nguyen2020PoisoningIoT} & 2020 & \makecell{D1} & -- & Black-box & \cmark & \cmark & \xmark & \xmark & \xmark & HFL & IoTD \\
GRA~\cite{Liu2020BackdoorAA} & 2020 & D2 & M3 & Full-box & \xmark & \cmark & \xmark & \xmark & \xmark & VFL & IC \\
Edge-case~\cite{wang2020attack} & 2020 & \makecell{D1} & -- & Black-box & \xmark & \cmark & \xmark & \xmark & \cmark & HFL & \makecell[c]{IC, NLP}                                                                     \\
AnaFL~\cite{Bhagoji2019AnalyzingFL} & 2019 & -- & M1, M2 & White-box & \xmark & \cmark & \cmark & \xmark & \cmark & HFL & \makecell{IC, LR}                                                                    \\
ALIE~\cite{ALittleIsEnough2019} & 2019 & -- & M2 & White-box & \cmark & \cmark & \xmark & \xmark & \cmark & HFL & IC                                                                                      \\
PGD~\cite{Sun2019CanYR} & 2019 & -- & M2 & White-box & \cmark & \cmark & \xmark & \xmark & \cmark & HFL & IC                                                                         \\
\makecell[l]{Constrain-and-scale}\cite{HTBD2018} & 2018 & D1, D2 & M1, M2 & White-box & \xmark &  \xmark & \cmark & \xmark & \cmark & HFL & \makecell{IC, NLP}                               \\
\makecell[l]{Model replacement} \cite{HTBD2018} & 2018 & D1, D2 & M1 & White-box & \xmark & \xmark & \cmark & \cmark & \cmark & HFL & \makecell{IC, NLP}                                       \\
Sybils~\cite{fung2018mitigating} & 2018 & \makecell{D1} & -- & Black-box & \cmark & \cmark & \xmark & \xmark & \xmark & HFL & Classification \\
\hline
\multicolumn{2}{l}{\cmark: YES/Applicable} &
\multicolumn{2}{l}{\xmark: NO/Not Applicable} &
\multicolumn{2}{l}{--: Not Main Focus} &
\multicolumn{2}{l}{D1: Semantic} & 
\multicolumn{3}{l}{D2: Artificial}    \\   

\multicolumn{2}{l}{M1: Scaling-based} & 
\multicolumn{2}{l}{M2: Constrain-based} & \multicolumn{2}{l}{M3: Gradient-replacement} & \multicolumn{2}{l}{M4: Partially Poisoning}   \\ 



\multicolumn{2}{l}{IC: Image Classification} &  \multicolumn{2}{l}{IoTD: IoT System} &
\multicolumn{2}{l}{LR: Logistic Regression} &
\multicolumn{2}{l}{NLP: Natural Language Processing} &
\end{tabular}
}
\end{table*}


\section{Techniques of Backdoor Attacks in FL}
\label{sec:backdoor-techniques}

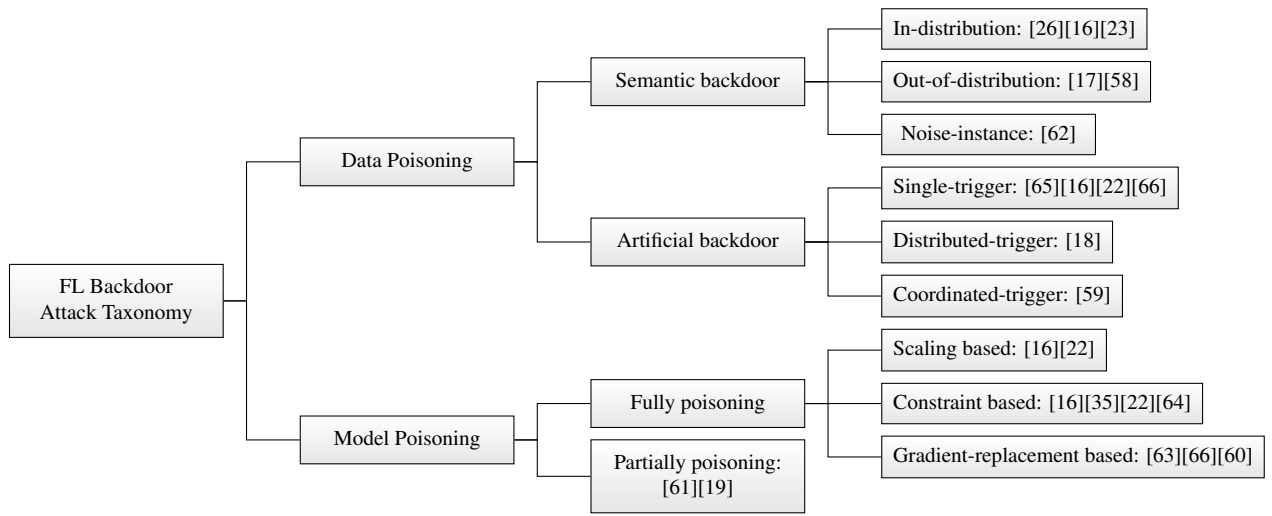
\begin{figure}[hbt!]
    \centering
    \scriptsize
    \begin{forest}
      for tree={
            inner sep=0.15cm, 
            l sep=10mm,
            s sep=1.5mm,
            minimum width=10em,
    top color=white, bottom color=gray!20,
            grow'=0, draw, thin,
            child anchor=west,parent anchor=east,
            anchor=west,calign=center,
      edge path={
            \noexpand\path[\forestoption{edge}]
              (!u.parent anchor) --  +(3mm,0mm) |- (.child anchor)\forestoption{edge label};
            },
      }
        [\makecell{FL Backdoor\\Attack Taxonomy}
            [\makecell{Data Poisoning}
                [\makecell{Semantic backdoor}
                    [In-distribution: \cite{fung2018mitigating}\cite{bagdasaryan2020backdoor}\cite{Nguyen2020PoisoningIoT}]
                    [Out-of-distribution: \cite{wang2020attack}\cite{rare-Yoo2022BackdoorAI}] 
                    [Noise-instance: \cite{PoisonGAN}]]
                [\makecell{Artificial backdoor}
                    [Single-trigger: \cite{Gu2019BadNetsEB}\cite{bagdasaryan2020backdoor}\cite{Sun2019CanYR}\cite{Liu2021BatchLI-VFL}]
                    [Distributed-trigger: \cite{xie2020dba}]
                    [Coordinated-trigger: \cite{Gong2022CoordinatedBA}]
                ]
            ]
            [\makecell{Model Poisoning}
                [\makecell{Fully poisoning}
                        [Scaling based: \cite{bagdasaryan2020backdoor}\cite{Sun2019CanYR}]
                        [Constraint based: \cite{bagdasaryan2020backdoor}\cite{Bhagoji2019AnalyzingFL}\cite{Sun2019CanYR}\cite{ALittleIsEnough2019}]
                        [Gradient-replacement based: \cite{Liu2020BackdoorAA}\cite{Liu2021BatchLI-VFL}\cite{Zou2022VFL}]
                ]
                [\makecell{Partially poisoning:\\ \cite{Zhou2021DeepMP}\cite{Neurotoxin}}]
            ]
        ]
    \end{forest}
    \caption{Taxonomy of FL \tuan{Backdoor Attacks.}}
    \label{fig:FL-attack-tao}
\end{figure}

The backdoor attack is first introduced in FL by Bagdasaryan et al.~\cite{bagdasaryan2020backdoor}. Since then, backdoor attacks have received widespread attention and became the primary security threat in FL. 
In most existing works~\cite{wang2020attack, Neurotoxin, Nguyen2022-FLAME, 2021-AdvancesOPFL, RODRIGUEZBARROSO2022-FL-threat-survey}, backdoor attacks are often conducted in both local training stages: training data collection and local training procedures. 
The goal of the adversary during the former stage is to manipulate a poisoned training dataset in order to corrupt the corresponding local model (i.e., data poisoning attacks). After that, the adversary alters the poisoned model to enhance the attack effectiveness and this is referred to as model poisoning attacks.
In this section, we investigate different techniques to manipulate the above-mentioned data poisoning and model poisoning attacks, as shown in Figure~\ref{fig:FL-attack-tao}. We then discuss how the adversary combines these techniques and compares their state-of-the-art backdoor attacks from perspectives of adversary assumption and attack efficiency in Table~\ref{tab:backdoor-attack-comparision}.

\subsection{Techniques for Data Poisoning Attacks}
\begin{figure}[tbh]
     \centering
     \includegraphics[width=0.95\textwidth]{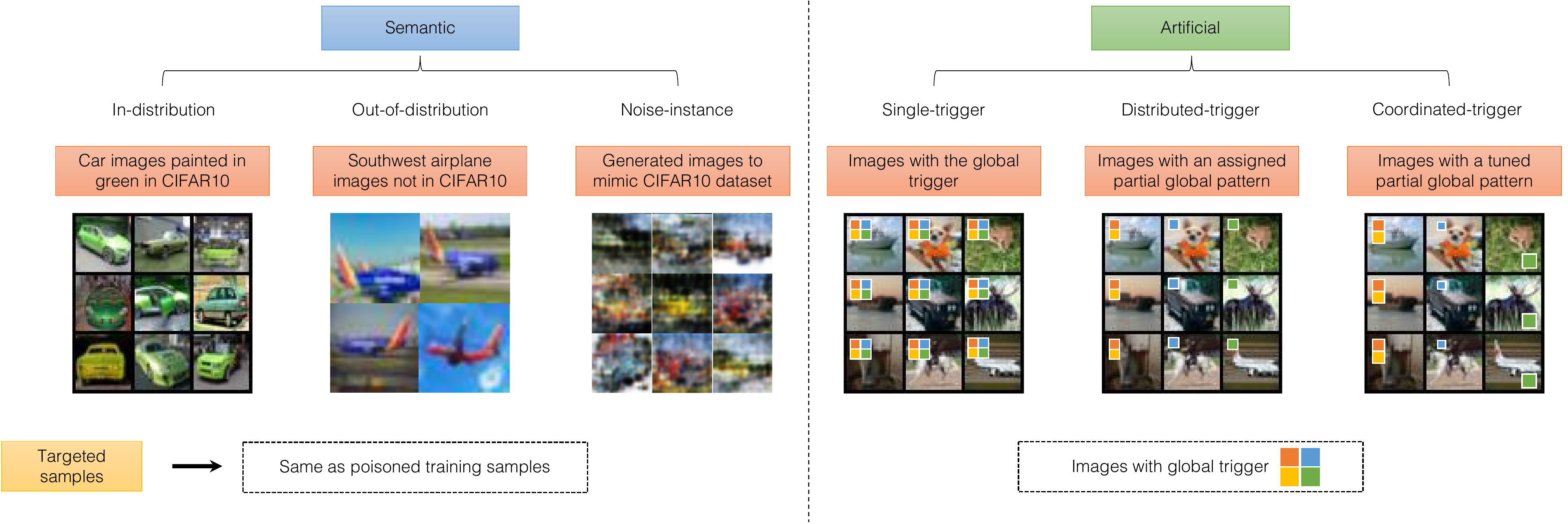}
     \caption{\footnotesize Illustration of poisoned training samples representative for each data poisoning technique. To backdoor a CIFAR-10 classifier: (In-distribution) green car images from CIFAR-10~\cite{CIFAR-10} labeled as ``bird"; (Out-of-distribution) southwest airplanes not from CIFAR-10 labeled as ``truck"; (Noise-instance) generated images from GAN model mimicking CIFAR-10 dataset. To backdoor model with a global trigger: (Single-trigger) all compromised clients insert global trigger to create poisoned images; (Distributed-trigger) each malicious client is assigned a partial global trigger (local trigger); (Coordinated trigger) each malicious client is assigned a local trigger and learns the optimal values for it.}
     \label{fig:data_poisoning_examples}
 \end{figure}

In data poisoning attacks, it is assumed that the adversary has complete control over the training data collection process of compromised clients. 
Most of the time, the poisoned training dataset has clean and poisoned samples with a backdoor trigger. As a result, the fundamental research topic in this subsection is how to generate backdoored samples.
Regarding the characteristics of backdoored samples, data poisoning attacks can be further classified into semantic backdoor attack and artificial backdoor attack. In semantic backdoor attacks, the targeted inputs should have specific properties, e.g., a pixel pattern or a word sequence, e.g., cars with striped pattern~\cite{bagdasaryan2020backdoor}. 
In this category of attack, no modification is conducted to modify the features of backdoored samples. On the other hand, artificial backdoor attacks~\cite{bagdasaryan2020backdoor, xie2020dba, Gong2022CoordinatedBA} aim to misclassify any poisoned input containing a backdoor trigger. Note that, these backdoored samples are created by artificially inserting triggers into the clean inputs. In the testing phase, a semantic backdoor attack can prompt misbehavior without any modification on the input samples while the artificial backdoor attack needs additional interference to manipulate targeted samples. We illustrated different techniques to manipulate poisoned training data in Figure~\ref{fig:data_poisoning_examples}.
\subsubsection{Semantic Backdoor Attacks}
In semantic backdoor attacks, the adversary poisons benign samples from compromised clients by flipping their labels. There are various policies in this strategy for selecting benign samples to poison. In particular, \cite{fung2018mitigating,SemiTargetedMP-Sun2022,tolpegin2020data} target the samples belonging to the global distribution, and these samples may be a part of other participants' training data or the testing set that the orchestration server may hold. This approach is referred to as in-distribution backdoor attacks.
For instance, all images of class ``1" are labeled as ``0" in~\cite{fung2018mitigating} and images of ``dog" are flipped to ``cat" in~\cite{tolpegin2020data}. On the other hand, in~\cite{bagdasaryan2020backdoor}, the attacker specifically targets samples possessing particular characteristics such as the unusual car color (e.g., green), the presence of a special object in the scene (e.g., stripped pattern) and trigger sentence ends with an attacker-chosen target word in word prediction problem. ~\cite{Nguyen2020PoisoningIoT} proposed an attack scenario to backdoor an FL-based IoT Intrusion Detection System in which the adversary targets packet sequences of malicious traffic from specific malware (e.g., IoT malware like Mirai malware). Nevertheless, the biggest limitation of these methods is that the updates from benign clients may dilute the backdoor effect. 



Recognizing the limitations of the previous works, Wang and Yoo~\cite{wang2020attack,rare-Yoo2022BackdoorAI} chose out-of-distribution samples that were far from the global distribution and unlikely to appear on the validation set of training sets of benign clients to backdoor the model. The key idea behind the success of these attacks is that the targeted set samples frequently lie at the tail of the data distribution of the benign clients, ensuring that the impact of backdoors is not easily diluted. 
Specifically, the author in~\cite{wang2020attack} proposed an edge-case backdoor attack in which the adversary targeted the edge-case samples (e.g., Southwest airplane images), which are not likely to appear in the benign clients' training data, that is, they are located in the tail of the input distribution (e.g., CIFAR-10~\cite{CIFAR-10}). Besides, authors in~\cite{rare-Yoo2022BackdoorAI} proposed ultimate rare word embedding to trigger the backdoor in \tuan{NLP} domain. The efficacy of this strategy is shown in Table~\ref{tab:backdoor-attack-comparision}, where edge-case backdoor attacks can perform successfully even with only one client and no model poisoning.

These methods mentioned above often require some knowledge of the target model, such as a portion of global data distribution, which turns out unpractical under specific scenarios. Different from the two approaches mentioned above,~\citep{PoisonGAN} proposed to train a GAN network during the local training process and employ the shared model to generate crafted samples and leverage these samples to backdoor the model. Since the adversary may not be knowledgeable about the data distribution of benign clients, so leveraging the GAN network to mimic other participants’ training samples helps the attack conduct a backdoor attack~\citep{PoisonGAN} under such a limited adversary's capability. In this case, the backdoored sample is the noise instances generated by the GAN network.

\subsubsection{Artificial Backdoor Attacks}
\dung{In contrast to semantic backdoor attacks, the targeted samples do not have to share specified properties and can belong to various classes. In addition, the adversary needs to artificially poison benign samples before flipping their labels. In other words, the ``key" for the backdoor does not naturally exist in the samples (i.e., the adversary adds pattern ``L" into the corner of images to activate the backdoor).}
The key idea of this strategy of attack is to poison a model such that in the presence of a backdoor trigger, the model will misbehave while maintaining normal behaviors in the absence of a trigger. This strategy is aligned with ``digital attack" in ML, in which the adversary digitally inserts a random pixel block into an input~\citep{Wenger2021PhysicalBA,Li2022BackdoorLA}. Due to the decentralized characteristics of FL, the different manners to distribute the trigger result in different attacking methods.

    Existing backdoor attacks against FL are mainly based on a single trigger, that is, all the compromised clients inject the same trigger into their local training dataset~\citep{bagdasaryan2020backdoor, Sun2019CanYR, Gu2019BadNetsEB, Liu2021BatchLI-VFL}. The trigger used in this approach is often set randomly and determinedly (e.g., square, cross patterns at the redundant pixels of images). At the inference process, the inserted trigger(s) to malicious clients are employed to trigger the aggregated model. Although the effectiveness of the backdoor inserted is proved to be significant~\citep{bagdasaryan2020backdoor}, the above works have not fully exploited the decentralized nature of the FL as they embedded the same trigger(s) to all adversarial clients (cf.~\citep{xie2020dba}).
    
    Observing the shortcomings of the previous regime,~\cite{xie2020dba} proposed distributed backdoor attack (DBA), which decomposes the objective trigger into many local triggers and assigns them to the corresponding compromised participants. In particular, each adversarial party uses its local trigger to poison the training data and sends the poisoned update to the server after it has finished local training. Unlike the previous technique, the attacker constructs a global trigger by combining local triggers rather than using them individually to activate the backdoor, and we refer to this attack technique as a distributed-trigger backdoor attack. Even though the global model wasn't present during training, DBA could still achieve a higher attack success rate and be more stealthy than a single-trigger attack strategy.
    
In prior techniques, the adversary's chosen trigger is frequently produced independently of the learning model and the learning procedure (e.g., a logo, a sticker, or a pixel perturbation). Therefore, such backdoor attacks do not fully exploit the collaboration between multiple malicious users during the training phase~\citep{Gong2022CoordinatedBA}. To address this shortage,~\cite{Gong2022CoordinatedBA} newly introduced coordinated-trigger backdoor attack, in which the adversary leverages a model-dependent trigger to inject the backdoor more efficiently. The model-dependent trigger is the optimal trigger configuration for each malicious participant. This is accomplished using a sub-training process that seeks the ideal value assignment of the trigger in terms of shape, size, and placement.
After the local trigger is generated for each adversarial party, the local training dataset will be poisoned based on the trigger. At the inference step, the global trigger is constructed by combining local triggers, this idea is analogous to~\citep{xie2020dba}. To this end, the model-dependent trigger is proven more efficient than the standard random trigger in previous works.
\subsection{Techniques for Model Poisoning Attacks}
In FL, even data poisoning directly results in poisoned updates, which are then aggregated to the global model, it is rarely used as a stand-alone backdoor attack strategy. The reason is that the aggregation cancels out most of the backdoored model’s contribution, and then the global model quickly forgets the backdoor~\cite{bagdasaryan2020backdoor,Neurotoxin,Gong2022CoordinatedBA}.
As a result, many works proposed combining data poisoning and model poisoning techniques
to enhance the effect of a backdoor attack. This strategy requires that the adversary have complete control over the training procedure and the hyperparameters (e.g., number of epochs and learning rate) and be free to modify the model parameters before submitting it~\cite{bagdasaryan2020backdoor}. This approach demonstrates its efficiency 
in various scenarios in the literature~\cite{bagdasaryan2020backdoor, wang2020attack, rare-Yoo2022BackdoorAI, Liu2021BatchLI-VFL}. Based on the range of poisoned parts in model parameters, we can categorize existing works into \textit{Fully poisoning attack} and \textit{Partially poisoning attack} as followings.
\subsubsection{Fully Poisoning Attacks }
Because the average approach is the most frequent way of aggregating local updates from clients, the most simplistic way to amplify the backdoor effect is to scale the updates from adversarial clients to dominate the updates from benign ones.
\citep{bagdasaryan2020backdoor} first introduced the model replacement method, in which the attacker attempts to replace the new global model with the poisoned model by scaling the poisoned update by a wisely-chosen factor. This strategy necessitates a careful assessment of global parameters and performs better when the global model is nearing convergence~\citep{bagdasaryan2020backdoor}.  This technique is widely employed in subsequent works and illustrates its effectiveness in intensifying the backdoor~\citep{Sun2019CanYR,wang2020attack}. However, given the range of FL defenses using clipping and restricting methods, straight scaling appears to be naive to success.
    
For stealthier model poisoning attacks, the attacker restricts the local updates from malicious clients so that the server's anomaly detector doesn't notice them. This is done by considering feasible anomaly detectors which may be used.~\citep{bagdasaryan2020backdoor,Bhagoji2019AnalyzingFL} proposed to modify the objective (loss) function by adding anomaly detection terms. The terms considered are formulated from the assumptions of any anomaly detection (e.g., the p-norm distance between weight matrices, validation accuracy). In~\citep{Sun2019CanYR,wang2020attack}, the projected gradient descent (PGD) attack is introduced to be more resistant to many defense mechanisms. In a PGD attack, the attacker projects their model on a small ball centered around the previous iteration's global model. This is performed so that the attacker's model doesn't change much from the global model at each FL round. Along with the line,~\citep{ALittleIsEnough2019} established a method to calculate a perturbation range in which the attacker can change the parameters without being detected even in \tuan{Independent and Identically Distributed (IID)} settings. From this perturbation range, an additional clipping step is conducted to better cover the malicious updates.
        
        
        The model poisoning attack strategies mentioned above originate from the design of Horizontal FL, wherein the participating parties own the labels of their data training samples. However, to the best of our knowledge, these techniques have not been verified or fully investigated in the Vertical FL scheme. Due to this fact,~\citep{Liu2020BackdoorAA,Liu2021BatchLI-VFL} introduced \textit{Gradient-replacement backdoor attack}, which is applicable to VFL even when the adversary owns only one clean sample belonging to the targeted class. Specifically, the attacker in~\citep{Liu2020BackdoorAA} records the intermediate gradients of clean samples of the targeted class and replaces the gradients of poisoned samples with these and uses these poisoned gradients to update the model.~\citep{Zou2022VFL} shown that even with HE-protected communication, the backdoor attack can also be conducted by directly replacing encrypted communicated messages without decryption using gradient replacement method. 

\subsubsection{Partially Poisoning Attacks }
Unlike the previous direction, which is fully poisoning the model parameters of the malicious clients,~\citep{Zhou2021DeepMP} demonstrated that the backdoor insertion could be conducted effectively without fully poisoning the whole space of model parameters. Specifically, they proposed an optimization-based model poisoning attack that injects adversarial neurons in the redundant space of a neural network to keep the stealth and persistence of an attack. To determine the redundant space, the Hessian matrix is leveraged to measure the distance and direction (i.e., “important”) of the update for the main task for each neuron. Then, an additional term is added to the loss function to avoid injecting poisoning neurons in positions that are particularly relevant to the main task. More recently,~\citep{Neurotoxin} proposed Neurotoxin, wherein the adversary employs the coordinates that the benign agents are unlikely
to update to implant the backdoored model to prolong the durability of the backdoor. In Neurotoxin, instead of directly updating the poisoned model by gradient computed on poisoning data, the attacker projects gradient onto coordinate-wise constraint, the bottom$-k\%$ coordinates of
the observed, benign gradient. The common objective of partially poisoning attacks is to prevent catastrophic forgetting of the adversarial task and prolong the durability of the backdoor's impact. 
\subsection{Comparison of FL Backdoor Attacks}
We first compare the existing attacks in the following ten dimensions belonging to three main aspects: backdoor characteristics, adversary assumptions, and attack efficiency in Table~\ref{tab:backdoor-attack-comparision}.

\textbf{Backdoor Characteristics. }Although data poisoning attacks result in poisoned model updates that are then aggregated into the global model, the majority of cutting-edge attacks combine data poisoning with model poisoning to enhance the backdoor effect.\\
--\textit{ Data Poisoning Techniques: }Following~\cite{HTBD2018}'s introduction of two approaches for conducting data poisoning attacks: artificial and semantic ones, further research aimed at developing more sophisticated attacks followed either direction. For instance, PoisonGAN~\cite{PoisonGAN} and CBA~\cite{Gong2022CoordinatedBA} are two significant advancements corresponding to semantic and artificial backdoor attacks, respectively.\\
--\textit{ Model Poisoning Techniques: } At the beginning stage of backdoor attacks in FL, scaling and constraining-based techniques are commonly used~\cite{HTBD2018, Sun2019CanYR, Bhagoji2019AnalyzingFL, ALittleIsEnough2019} to intensify the backdoor effect and cover anomaly of poisoned updates. More recently, adversaries exploit sparse characteristics of neural networks to conduct partially poisoning models~\cite{Neurotoxin, rare-Yoo2022BackdoorAI, Zhou2021DeepMP}. On the other hand, authors in~\cite{Zou2022VFL, Liu2020BackdoorAA} made the first attempts to implant backdoors in VFL by using the gradient-replacement technique to manipulate poisoned updates caused by artificially poisoned samples.\\
--\textit{ Accessibility: }According to Table~\ref{tab:backdoor-attack-comparision}, the black-box attack is rarely applied as a stand-alone strategy, despite being the simplest approach for inserting a backdoor. As presented, only~\cite{wang2020attack, fung2018mitigating} can be applied as a black-box attack, while the remaining attack approaches leverage white-box attack to facilitate model poisoning techniques.

\textbf{Adversary Assumptions.} Existing attack strategies are designed with specific adversary assumptions in consideration, and three major assumptions can be summarized as follows: the number of compromised participants, the frequency of attacks, and the convergence stage constraint to implant a backdoor. To ensure a successful attack, corresponding assumptions must hold true, which implies that any attack technique is practical. \\
--\textit{ Collusion Required: }Many methods require participant collusion, so these strategies are only applicable under favorable conditions, i.e., the adversary controls sufficient compromised clients~\cite{xie2020dba, Sun2019CanYR, Nguyen2020PoisoningIoT, fung2018mitigating, rare-Yoo2022BackdoorAI, Gong2022CoordinatedBA, PoisonGAN, ALittleIsEnough2019}. However, in large-scale FL systems, this condition is difficult to be satisfied.
The remaining methods not requiring participant collusion are often combined with other additional model poisoning attacks to strengthen the backdoor effect of one malicious client. Unlike previous methods, edge-case~\cite{wang2020attack} demonstrates its efficiency even when the adversary controls only one client and does not employ any model poisoning techniques.\\
--\textit{ Continuous Attack: }We can see apart from~\cite{xie2020dba, HTBD2018, Zhou2021DeepMP}, existing backdoor attacks are continuous attacks, in which the malicious clients participate in the training for multiple rounds. This continually reminds the global model about the backdoor task, which can reduce the backdoor dilution phenomenon caused by benign updates. Otherwise, the methods in~\cite{xie2020dba, HTBD2018, Zhou2021DeepMP} can be employed as single-shot attacks, in which the adversary can inject the backdoor in only one round. This attack strategy is more preferable, especially in a large-scale FL system, where the participant probability of each client is relatively small. \\
--\textit{ Convergence Stage Constraint:} The efficiency of single-shot attacks depends on the period that the backdoor is inserted. Certainly, apart from~\cite{Zhou2021DeepMP}, other single-shot attacks~\cite{xie2020dba, HTBD2018} are only effective when the global model is close to convergence. Although the adversary can employ recent methods to estimate the next global model~\cite{Bhagoji2019AnalyzingFL} or facilitate the convergence of global model~\cite{Liu2022TechnicalRA}, these methods require substantial complicated technical skills and knowledge about global distribution.

\textbf{Backdoor Efficiency. }\\
--\textit{ Extended Durability: }One challenge to backdoor attack designing is that the malicious clients often account for just a small portion of total clients in reality, i.e., $\left[0.01, 1\right]\%$ (cf.~\citep{Shejwalkar2021BackTT}). Therefore, the poisoned updates may be easily diluted by the benign updates, which is also known as ``catastrophic forget" in machine learning. Although model-replacement attack~\cite{HTBD2018} can extend the backdoor longevity, it was not until 2021 that~\citep{Neurotoxin,Zhou2021DeepMP} officially consider durability as an attack objective. To achieve the goal, partial model poisoning attacks are employed to prolong backdoor durability, and this opens a new novelty to designing a robust and durable backdoor attack. This strategy exploits the sparse nature of gradients in stochastic gradient descent (SGD) and poisons only a subset of neurons while preserving the remaining neurons unaffected.\\
--\textit{ Stealthiness Consideration: }The emergence of defending mechanisms has challenged FL adversaries. This prompted more works to consider the stealthiness of their backdoor attacks. Constraint-based model poisoning and partially-poisoning attacks are two mainstream approaches for achieving this goal~\cite{bagdasaryan2020backdoor, Neurotoxin, Sun2019CanYR, Bhagoji2019AnalyzingFL, rare-Yoo2022BackdoorAI, Zhou2021DeepMP, ALittleIsEnough2019}, and constraint-based methods are more popular. Although these methods can bypass common defenses, the adversary must be knowledgeable of difficult-to-achieve information in the physical world such as the aggregation operator~\cite{bagdasaryan2020backdoor, Zhou2021DeepMP}, global data, and employed defenses~\cite{Sun2019CanYR,Bhagoji2019AnalyzingFL}.\\
-- \textit{FL Type: } Most existing works focus on HFL, in which there is that the aggregation server is honest and there are one to several malicious clients, which are totally controlled by adversaries. There are only~\cite{Zou2022VFL, Liu2020BackdoorAA} proposed backdoor attacks in VFL with gradient-replacement techniques although VFL provides many favorable conditions to conduct backdoor attacks. For example, VFL is often involved by a much less number of participants in HFL, i.e., less than five~\cite{10.1145/3298981}, and each participant in VFL possesses a part of a global model. To the best of our knowledge, backdoor attacks have not appeared in FTL.\\
-- \textit{Applications: }Backdoor attacks have been evaluated under several domains in FL including image classification, IoT, and natural language processing. We can see that most attacks target image classification. To tailor backdoor attacks for a specific domain, i.e., network intrusion detection for IoT~\cite{Nguyen2020PoisoningIoT}, the adversary needs to develop a specialized data poisoning strategy.

\section{Backdoor Defense Methodologies}
\label{sec:backdoor_defense_methodologies}
\tuan{In the literature, there are different strategies applicable to handle backdoor attacks in FL, with some specifically designed for this type of attack (dedicated), while others aim to defend against multiple attack types, including backdoor attacks (non-dedicated)}. These defenses can be implemented at different stages of the FL training process, resulting in various methods and approaches. 
For instance, server-side defenses are predicated on the assumption that the orchestration server can be trusted as a collector and aggregator of local updates from clients. In contrast, client-side defenses aim to protect the robustness of FL when the trustworthiness of the server cannot be assumed. While some strategies were specifically designed for FL backdoor attacks, others, such as Krum~\cite{blanchard2017machine} and geometric mean~\cite{RFA2022} for mitigating Byzantine attacks, have also been effective in defending against such attacks despite having strong assumptions (e.g., IID data) and with specific limitations. 

\begin{figure*}[tbh]
     \centering
     \includegraphics[width=0.95\linewidth]{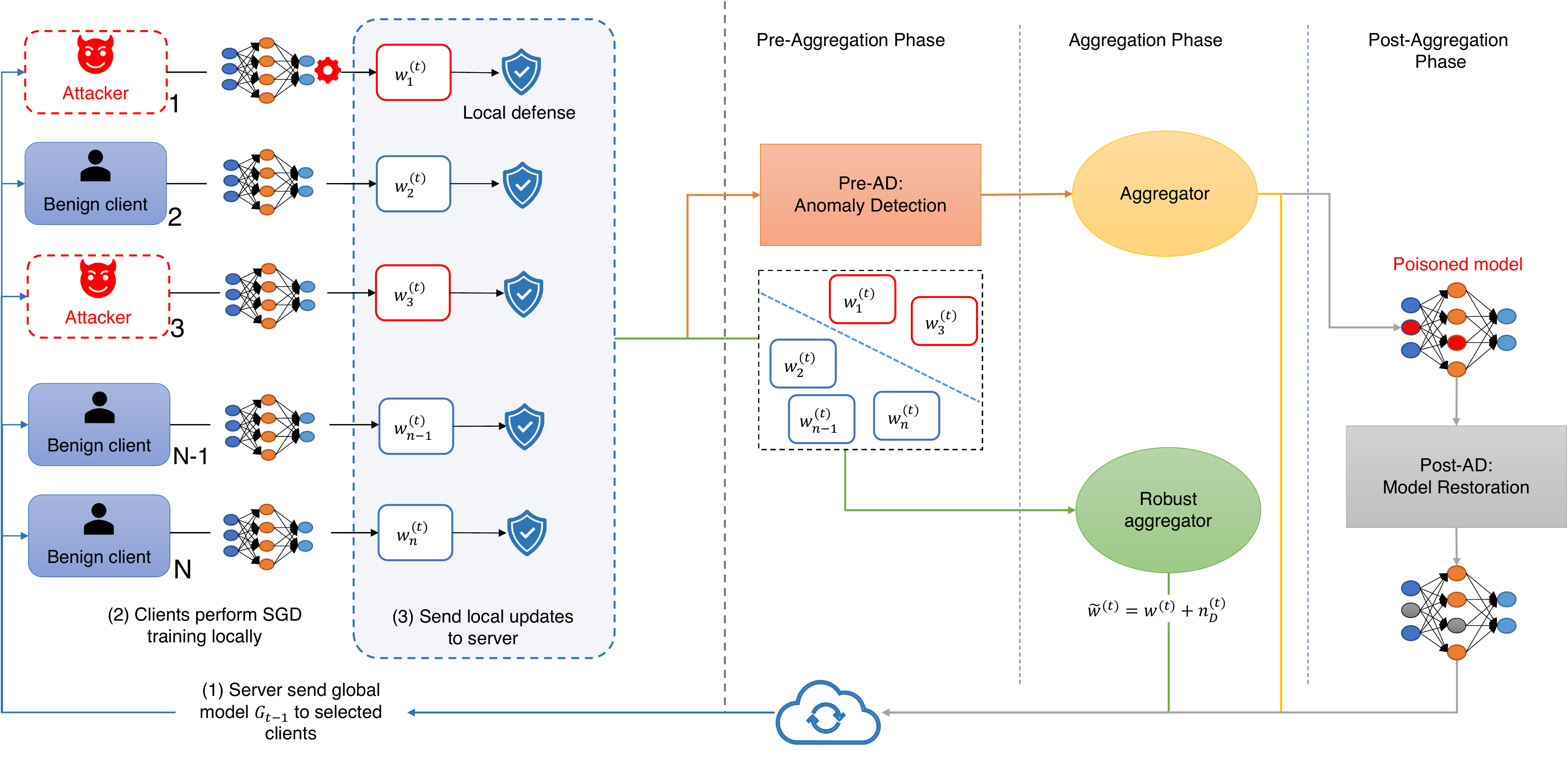}
     \caption{Overview of different categories of backdoor defenses in FL: previous-aggregation defense (Pre-AD), in-aggregation defense (In-AD), and post-aggregation defense (Post-AD).}
     \label{fig:defense_pipeline}
 \end{figure*}

\begin{figure}[tbh]
\centering
\scriptsize
\begin{forest}
      for tree={
            inner sep=0.15cm, 
            l sep=10mm,
            s sep=1.5mm,
            minimum width=10em,
    top color=white, bottom color=gray!20,
            grow'=0, draw, thin,
            child anchor=west,parent anchor=east,
            anchor=west,calign=center,
      edge path={
            \noexpand\path[\forestoption{edge}]
              (!u.parent anchor) --  +(3mm,0mm) |- (.child anchor)\forestoption{edge label};
            },
      }
    [\makecell{FL Backdoor\\Defense Taxonomy}
        [Pre-AD
            [Graph-based:~\cite{cao2019understanding}]
            [Clustering:~\cite{blanchard2017machine, munoz2019byzantine, shen2016auror, tolpegin2020data, HTBD2018, Bhagoji2019AnalyzingFL, fung2018mitigating, sattler2020byzantine, preuveneers2018chained, nguyen2019diot, li2020learning, blanchard2017machine, Nguyen2022-FLAME}]
        ]
        [In-AD
            [Robust Aggregation Rule:~\cite{yin2018byzantine, guerraoui2018hidden, pmlr-v80-yin18a, ozdayi2021defending, bernstein2018signsgd}]
            [\makecell{Model Smoothness \& Perturbation}:~\cite{Nguyen2022-FLAME, rieger2022deepsight, xie2021crfl, sun2021fl}]
            [Differential Privacy:~\cite{Sun2019CanYR, naseri2020toward}]
            [Local validation:~\cite{andreina2021baffle}]
        ]
        [Post-AD
            [Pruning Neurons:~\cite{Wu2020MitigatingBA}]
            [Unlearning:~\cite{wu2022federated}]
        ]]
\end{forest}
\caption{Taxonomy of FL backdoor defense.}
\label{fig:FL-defense-tao}
\end{figure}
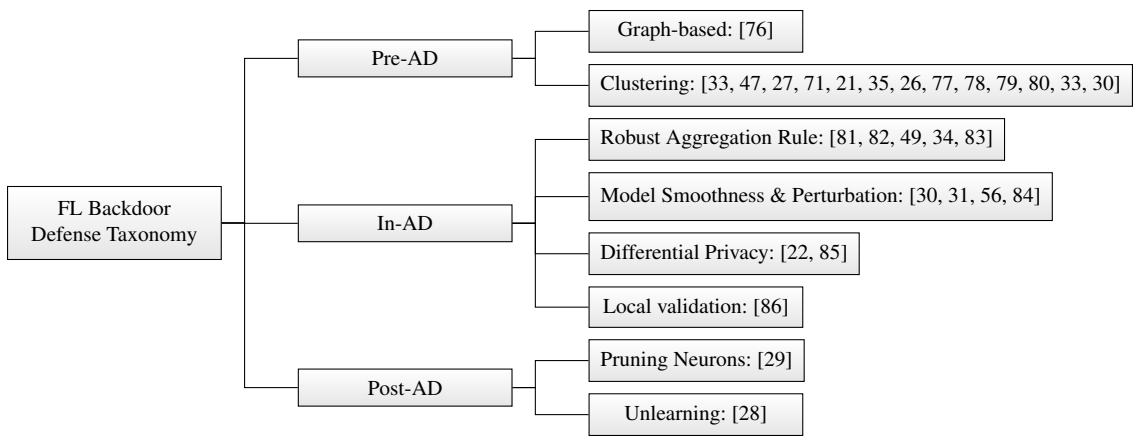

In general, the FL backdoor defenses can be grouped into three categories based on different methodologies: previous-aggregation defense (Pre-AD), which uses anomaly detection techniques; in-aggregation defense (In-AD), which relies on robust training techniques; and post-aggregation defense (Post-AD), which involves model restoration. We give the overview of these defenses in Figure~\ref{fig:defense_pipeline} and the taxonomy of each defense in Figure~\ref{fig:FL-defense-tao}.
 
\subsection{Previous-aggregation Defenses}
Pre-AD methods are implemented before the server aggregates model updates from clients. These methods first identify adversarial clients as anomalous data in the distribution of local model updates and then exclude them from the aggregation. Specifically, the Pre-AD methods rely on the assumption that malicious client model updates are similar and use either unsupervised or supervised ML techniques to differentiate between benign and malicious updates. Examples include Krum~\cite{blanchard2017machine}, AFA~\cite{munoz2019byzantine}, and Auror~\cite{shen2016auror}, which use distance measurements such as the \tuan{Mahalanobis Distance~\cite{mahalanobis1936generalised} and Cosine Similarity~\cite{singhal2001modern}} under the assumption of either IID or non-IID data distribution. However, model updates are often highly dimensional, making it difficult to apply anomaly detection techniques effectively. To address this, some works use dimensional reduction techniques such as PCA to make the data more manageable \cite{tolpegin2020data}. These approaches typically rely on the Euclidean Distance for clustering, which can be vulnerable to stealthy attacks like constraint-based attacks~\cite{HTBD2018,Bhagoji2019AnalyzingFL}. In FoolsGold~\cite{fung2018mitigating}, the defense mechanism inspects client updates based on the similarity of their model updates, with the assumption that malicious updates will behave more similarly to benign updates.

Anomaly detection can be performed using ML techniques, such as clustering and graph-based methods. For example, in~\cite{sattler2020byzantine}, model updates are divided into clusters based on Cosine Distance, and in~\cite{preuveneers2018chained}, an unsupervised deep learning anomaly detection system is integrated into a blockchain process. Graph-based anomaly detection is proposed in~\cite{cao2019understanding}, where the authors build a graph of local model updates and identify benign models by solving a maximum clique problem. Anomaly-based systems based on Gated Recurrent Units (GRUs) have been tested on IoT-specific datasets in~\cite{nguyen2019diot}. Li et al.~\cite{li2020learning} \tuan{proposed} a spectral anomaly detection framework using a latent space and an encoder-decoder model. Malicious updates are identified as those that produce higher generation errors than benign ones. In~\cite{rieger2022deepsight}, the authors \tuan{proposed} DeepSight, a novel model filtering approach that characterizes the distribution of data used to train model updates and measures the differences in the internal structure and outputs of NNs to identify and eliminate model clusters containing poisoned models. The effectiveness of existing weight clipping-based defenses in mitigating the backdoor contributions of possibly undetected poisoned models is also demonstrated. In addition, FLDetector~\cite{zhang2022fldetector} \tuan{suggested} a method for detecting malicious clients by examining the consistency of their model updates. Essentially, the server predicts a client's model update in each iteration based on past updates, and if the received model update from the client differs significantly from the predicted update over multiple iterations, the client is flagged as malicious.

Defenses against malicious clients in FL can be vulnerable to certain attack scenarios and impose strong assumptions about the adversary's capabilities. Multi-krum~\cite{blanchard2017machine} fails to mitigate edge-case backdoor attacks~\cite{wang2020attack} in non-IID data distributions, and FoolGold~\cite{fung2018mitigating} is vulnerable to constrain-and-scale attacks~\cite{bagdasaryan2020backdoor}. To address this, Nguyen et al.~\cite{Nguyen2022-FLAME} studied multi-target backdoor attacks, which do not assume a fixed number of adversaries or data distribution. FLAME~\cite{Nguyen2022-FLAME} \tuan{used} the HDBSCAN algorithm to detect malicious updates and combines model filtering with poison elimination to detect and remove malicious updates and is robust against inference attacks. However, FLAME requires more computational resources than traditional FL processes. Li et al.~\cite{li2020learning}'s method is effective at detecting multi-trigger backdoor attacks while maintaining high predi
ction accuracy for the benign main task. 

There are two main approaches for addressing malicious clients in FL: total exclusion and impact reduction. The first approach removes poisoned updates from malicious clients before aggregating updates from all clients~\cite{Nguyen2022-FLAME, blanchard2017machine}, and is effective when the proportion of malicious clients is high. However, its effectiveness against multi-target backdoor attacks is unknown, and it relies on the assumption that malicious clients will behave similarly at each round or that benign clients will have similar data distributions, which may not hold in certain cases such as fixed-frequency attacks. The second approach reduces the impact of malicious clients on the aggregated model, such as decreasing the learning rate of suspicious clients in FoolsGold~\cite{fung2018mitigating}. There is a risk of incorrectly detecting anomalous updates in cases where these assumptions do not hold.

\subsection{In-aggregation Defenses}
 The In-AD mechanism for FL operates while the server is aggregating local models, using techniques such as differential privacy, robust learning rates, smoothness and perturbation, and local validation to mitigate the effects of backdoors.
 
 \textbf{\textit{Differential Privacy (DP). }}
 DP has been shown to be effective against backdoors~\cite{Sun2019CanYR, naseri2020toward}, but it can compromise model performance under data imbalance~\cite{2021-AdvancesOPFL, bagdasaryan2019differential}, which is common in federated learning. DP-FedAvg~\cite{mcmahan2017learning} (Central-DP) is a differentially private aggregation strategy that removes extreme values by clipping the norm of model updates and adding Gaussian noise, but the required amount of noise significantly reduces task accuracy. Sun et al.~\cite{Sun2019CanYR} proposed Weak-DP, which adds sufficient Gaussian noise to defeat backdoors and preserve task accuracy, but it is not effective against constrain-based backdoor attacks~\cite{wang2020attack}. Additionally, differential privacy-based defenses can potentially affect the benign performance of the global model, as the clipping factors also change the weights of benign model updates~\cite{bagdasaryan2020backdoor, wang2020attack}.

\textbf{\textit{Model Smoothness and Perturbation. }} Despite the lack of robustness certification in previous defense approaches, Xie et al.~\cite{xie2021crfl} proposed the first general defense framework, CRFL, for training certifiable robust FL models against backdoor attacks. CRFL employs cropping and smoothing of model parameters to control model smoothness and generate sample robustness certification against backdoor attacks with limited amplitude. The smoothness and perturbation method is also used as an additional component to limit the L2-norm of individual updates to improve defense performance~\cite{Nguyen2022-FLAME, rieger2022deepsight}. Additionally, the FL-WBC~\cite{sun2021fl} method \tuan{aimed} to identify vulnerable parameter spaces in FL and perturb them during client training. FL-WBC also provides robustness guarantees against backdoor attacks and convergence guarantees to FedAvg~\cite{mcmahan2017communication}. These developments demonstrate promising steps toward improving the robustness of FL against backdoor attacks. In FLARE~\cite{wang2022flare}, a trust evaluation method is presented that calculates a trust score for each model update based on the differences between all pairs of model updates in terms of their penultimate layer representations values. FLARE assumes that the majority of clients are trustworthy, and assigns trust scores to each model update in a way that updates far from the cluster of benign updates receive low scores. The model updates are then aggregated with their trust scores serving as weights, and the global model is updated accordingly.

\textbf{\textit{Robust Aggregation Rule. }} Several approaches have been proposed to address the vulnerability of standard aggregation methods, such as FedAvg~\cite{mcmahan2017communication}, to backdoor attacks. For example, the use of the geometric median of local parameters as the global model has been proposed in~\cite{yin2018byzantine, guerraoui2018hidden}. Another approach is the use of the Median and $\alpha$-trimmed mean, which \tuan{replaced} the arithmetic mean with the median of model updates to increase robustness against attacks~\cite{pmlr-v80-yin18a}. Additionally, Ozdayi et al.~\cite{ozdayi2021defending} \tuan{proposed} the use of a Robust Learning Rate (RLR) as an improvement of signSGD~\cite{bernstein2018signsgd}, which adjusts the server's learning rate based on the agreement of client updates. 
Chen et al.~\cite{chen2020backdoor} introduced a defense mechanism inspired by matching networks, where the class of input is predicted based on its similarity with a support set of labeled examples. By removing the decision logic from the shared model, the success and persistence of backdoor attacks were greatly reduced.

\textbf{\textit{Local validation. }} BaFFle~\cite{andreina2021baffle} is a decentralized feedback-based mechanism that eliminates backdoors by using clients' data to validate the global model through a supernumerary validation process. Selected clients check the global model by calculating a validation function on secret data and report whether it is backdoored to the orchestration server. The server then decides whether to reject the global model based on the inconsistency of misclassification rates per class between the local model and the global model. The BaFFle is compliant with secure aggregation, but has limitations: it requires trigger data to activate the backdoor, does not work in non-IID data scenarios with a small number of clients, and is not effective against continuous attacks that corrupt FL training.


In-aggregation defenses, which are applicable in various FL schemes and preserve privacy, have little impact on the training process and are effective against artificial backdoor attacks~\cite{xie2020dba, Sun2019CanYR, zhang2020defending}. However, they primarily resist convergence attacks and do not completely discard poisoned local updates, allowing a significant percentage of compromised updates to impact the aggregated model. For example, the geometric median (RFA)~\cite{RFA2022} is vulnerable to distributed backdoor attacks~\cite{xie2020dba}, and RLR~\cite{ozdayi2021defending} can cause a trade-off between defense efficiency and performance on the main task. The effectiveness of these defenses and the trade-offs they incur under severe conditions such as a high ratio of malicious clients or non-IID data needs further evaluation.

It has been established that in a VFL scenario where features and models are partitioned among various parties, sample-level gradient information can be used to infer sensitive label information that should be kept confidential. To counter this issue, it is usual practice to encrypt sample-level messages with Homomorphic Encryption (HE) and only communicate batch-averaged local gradients among the parties.
However, Zou et al.~\cite{zou2022defending} showed that even with HE-protected communication, private labels can still be reconstructed with high accuracy via gradient inversion attacks, thereby challenging that batch-averaged information is secure to share under encryption. In response to this challenge,~\cite{zou2022defending} proposed a novel defense method, called Confusional Autoencoder (CAE), that utilizes autoencoder and entropy regularization techniques to conceal the true labels.

\subsection{Post-aggregation Defenses}
To ensure the integrity of the global model, a protective procedure is implemented after local models from clients, potentially including malicious ones, have been aggregated. The orchestration server subsequently reviews and amends the global model, maintaining valuable information and removing any corrupt updates from malicious clients.

Wu et al.~\cite{Wu2020MitigatingBA} introduced the first post-aggregation defense strategy for FL against backdoor attacks. Their approach involves identifying and removing neurons with low activation when presented with benign samples, as these neurons are likely to be dormant without the presence of the trigger. To address the issue of the server not having access to private training data, Wu et al.~\cite{Wu2020MitigatingBA} proposed a distributed pruning strategy. The server asks clients to record neuron activations using their local data and create a local pruning list, which is then used to determine a global pruning sequence. The server can adjust the pruning rate based on the current model's performance on a validation dataset and gather feedback from clients to finalize the pruning list.

Unlearning has recently gained attention in the field of ML~\cite{Bourtoule2021MachineU, Neel2021DescenttoDeleteGM, Gupta2021AdaptiveMU}, and its application to defend against backdoor attacks in FL has been explored by Wu et al.~\cite{wu2022federated}. Wu et al. demonstrated the use of Federated Unlearning for removing the effects of single-trigger backdoor attacks without significantly affecting overall performance (e.g., BA = 0\%). However, this method requires identifying malicious clients to be unlearned and has only been tested on artificial backdoor attacks, leaving its effectiveness against semantic backdoor attacks unknown.

\subsection{Comparing Approaches for Detecting and Mitigating Backdoor Attacks in FL}
We compare existing backdoor defenses in FL in terms of eight dimensions as shown in Table~\ref{tab:stat-defen}. The compared dimensions belong to three key perspectives of a backdoor defense: adversary assumptions, defensive requirements, and effectiveness.\\
\begin{table*}[hbt!]
\scriptsize
\centering
\caption{\tuan{A Comparison of the State-of-the-art Methods for Defending against Backdoor Attacks in FL}}
\vspace{1em}
\setlength{\tabcolsep}{0.6em}
\label{tab:stat-defen}

\resizebox{\textwidth}{!}{

\begin{tabular}{lllccccllc}
\hline
\multicolumn{1}{c}{\multirow{2}{*}{Categorization}} & \multicolumn{1}{c}{\multirow{2}{*}{Work}} & \multicolumn{3}{c}{Adversary assumptions} &
\multicolumn{2}{c}{Defensive Requirements}
& \multicolumn{2}{c}{Effectiveness } & \multicolumn{1}{c}{\multirow{2}{*}{Application}}\\

\cmidrule(lr){3-5} 
\cmidrule(lr){6-7}
\cmidrule(lr){8-9} 
\multicolumn{1}{c}{}   & \multicolumn{1}{c}{}  & \makecell { Defensive \\targets} & \makecell{Data\\distribution} & \makecell{\#Compromised\\(PMR)} & \makecell{Local update\\access} & \makecell{Model\\inference} & ASR             & MTA Change            \\
\hline
    \multirow{6}{*}{Pre-AD} 
    & FLDetector (2022)~\cite{zhang2022fldetector} & Backdoor Attacks & non-IID & $28\%$ & YES & NO  & $\le 2.4\%$ & $\pm 1.5$\% &\makecell{IC}\\ \cmidrule(l){2-10}
    & FLAME (2022)~\cite{nguyen2022flame} & Backdoor Attacks & non-IID & $ < 50\%$ & YES & NO  & $0\%$ & $\pm 0.5\%$ &\makecell{IoTD\\ IC/ NWP}\\ \cmidrule(l){2-10}
    & DeepSight (2022)~\cite{rieger2022deepsight} & Backdoor Attacks & non-IID & $ \le 45\%$ & YES & YES & $0\%$ & $\pm 0.5\%$ &\makecell{IoTD\\ IC/ NWP}\\ \cmidrule(l){2-10}
    & VAE (2020)~\cite{li2020learning} & \makecell[l]{In-distribution\\ Single-trigger} & non-IID &  $ \le 30\%$ & NO & NO & -- & -- & IC/ SA\\ \cmidrule(l){2-10} 
    & FoolsGold (2018)~\cite{fung2018mitigating} & In-distribution & non-IID & -- & YES & NO & $0\%$ & -- & IC\\ \cmidrule(l){2-10}
    & AUROR (2016)~\cite{shen2016auror} & In-distribution & IID & $ \le 30\%$ & YES & NO & $2\%$ & $\le 5\% $& IC \\
    \hline

    \multirow{8}{*}{In-AD} 
    & CAE (2022)~\cite{zou2022defending} & Gradient-replacement & non-IID & -- & NO & NO & -- & -- & IC \\ \cmidrule(l){2-10}
    & CRFL (2021)~\cite{xie2021crfl} & Distributed-trigger & non-IID &$\le 4\%$ & YES & NO & -- & -- & F\&B/ IC  \\ \cmidrule(l){2-10}
    
    \multicolumn{1}{c}{} & BaFFle (2021)~\cite{andreina2021baffle} & In-distribution & non-IID & -- & YES & YES & -- & -- & IC\\ \cmidrule(l){2-10}
    \multicolumn{1}{c}{} & RLR (2021)~\cite{ozdayi2021defending} & \makecell[l]{Distributed-trigger\\Single-trigger} & non-IID/ IID & 10\% & YES & NO & $\le 9\%$ & $< 5\%$& IC                   \\ \cmidrule(l){2-10}
    \multicolumn{1}{c}{} & DP (2020)~\cite{bagdasaryan2020backdoor} & Single-trigger & non-IID & $\le 5\%$ & YES & NO & -- & -- & IC/ NLP\\ \cmidrule(l){2-10}
    \multicolumn{1}{c}{} & Matching Networks (2020)~\cite{chen2020backdoor} & Single-trigger & IID & $25\% (1/ 4)$ & YES & NO & $\le 20\%$ & $+5\%$ & IC\\ \cmidrule(l){2-10}
    \multicolumn{1}{c}{} & FL-WBC (2020)~\cite{sun2021fl} & In-distribution & non-IID/ IID & $\le 50\%$ & YES & NO & -- & $\le 10\%$ & IC\\ \cmidrule(l){2-10}
    \multicolumn{1}{c}{} & Weak DP (2019)~\cite{Sun2019CanYR} & Single-trigger & non-IID & $3.33\%$ & YES & NO & -- & -- & IC\\ 
    \hline

    \multirow{1}{*}{Post-AD} 
    & KD Unlearning (2022)~\cite{wu2022federated} & Single-trigger & IID & $10\% (1/ 10)$ & YES & NO & $0\%$ & $\pm 1\%$ & IC\\ \cmidrule(l){2-10}
    & Pruning Neurons (2020)~\cite{Wu2020MitigatingBA} & Distributed-trigger & non-IID & $\le 10\%$ & YES & NO & $13\%$ & $ < 2\% $ & IC   \\
\hline
\multicolumn{2}{l}{Pre-AD: Previous-aggregation defense} & \multicolumn{2}{l}{In-AD: In-aggregation defense} & \multicolumn{2}{l}{Post-AD: Post-aggregation defense}\\
\multicolumn{2}{l}{NLP: Natural Language Processing} & \multicolumn{2}{l}{IoTD: IoT intrusion detection} & \multicolumn{2}{l}{IC: Image Classification} & \multicolumn{1}{l}{}\\
\multicolumn{2}{l}{SA: Sentiment Analysis} &  \multicolumn{2}{l}{B\&F: Banking and Finance} & \multicolumn{2}{l}{NWP: Next Word Prediction} & \multicolumn{4}{l}{IID: Independent and Identically Distributed}\\
\multicolumn{2}{l}{PMR: Poisoned Model Rate} & \multicolumn{2}{l}{ASR: Attack Success Rate} & \multicolumn{2}{l}{MTA: Main Task Accuracy} & \multicolumn{2}{l}{--: Not Available}\\

\end{tabular}}
\end{table*}
\dung{
\textbf{Adversary Assumptions. }Existing backdoor defenses in FL are based on specific observations and assumptions and often target specific types of backdoor attacks.\\
-- \textit{Defensive targets: }Most existing backdoor defenses are demonstrated to be efficient against in-distribution and single-trigger backdoor attacks. Recent Pre-AD defenses, i.e., FLAME~\cite{Nguyen2022-FLAME} and DeepSight~\cite{rieger2022deepsight}, are more versatile since they can handle various attack schemes. In fact, a robust backdoor defense should not rely on the type of backdoor attack.\\
-- \textit{Data distribution: }Except for~\cite{shen2016auror,wu2022federated}, most existing defenses are designed for the case in which all of the participants' training data adheres to non-IID. However, the data distribution
among participants in FL is often non-predictable. To be more applicable, the defenses should be effective under different data distributions, i.e., both IID and non-IID cases~\cite{ozdayi2021defending, sun2021fl}.\\
-- \textit{Poisoned Model Rate: }The Pre-AD methods can be employed when the PMR is sufficiently large (i.e., up to 50\%) because these methods aim at grouping the poisoned models into one group and the remaining group is benign. Other approaches, i.e., In-AD and Post-AD, are effective under smaller PMR such as less than 10\%.
}

\dung{
\textbf{Defensive Requirements. }Unlike ML, the orchestration server is not eligible to access the local training data, so the information that can be analyzed to defend against backdoor attacks is the local updates and their corresponding inference outputs.\\
-- \textit{Local Update Access: }Apart from~\cite{li2020learning, zou2022defending}, other defenses need to analyze all local model updates. This leads to an issue with computation overhead. Instead of examining entire model parameters, efficient methods such as last-layer parameter analysis can be utilized to circumvent this issue~\cite{rieger2022deepsight}.\\
-- \textit{Model inference: }To facilitate their defenses,~\cite{rieger2022deepsight, andreina2021baffle} need to consider inference results from local model updates. Although these strategies demonstrate efficiency in defending against backdoor attacks, this requires considerable computation costs. As a result, the remaining methods not requiring local model inferences are more relevant when the computation capacity of the central server is limited. 
}

\dung{
\textbf{Effectiveness. }Another issue with these defense methods is that they rely on too many assumptions about the
data distribution, number of clients participating, and number of attackers. This makes it difficult
to make a fair comparison between different approaches. \\
-- \textit{ASR: }The works ~\cite{Nguyen2022-FLAME, rieger2022deepsight} can mitigate ASR from 100\% to 0\% with a little change in main task accuracy, but they rely on a strong assumption about the number of attackers to make a distinction for malicious models. For example,~\cite{Nguyen2022-FLAME, rieger2022deepsight, li2020learning, sun2021fl} proposed defense methods that required a large percentage of malicious clients (up to 50\%) to be present in order to effectively detect and exclude them. This highlights the importance of understanding the specific threat model and the distribution of malicious clients in a given scenario. Therefore, it is uncertain how well these methods will perform in a realistic world.\\
-- \textit{MTA Change: }One issue with existing defenses in FL is the degradation of performance on the primary task. For example, methods such as ~\cite{shen2016auror, ozdayi2021defending, chen2020backdoor} result in a reduction of accuracy around 5\%. This underlines the importance of ongoing research and development of strong defense mechanisms to guarantee accurate and trustworthy model results.\\
-- \textit{Application: }Most applications of backdoor attacks have been implemented in \tuan{IC tasks~\citep{zhang2022fldetector, li2020learning, fung2018mitigating, shen2016auror, zou2022defending, andreina2021baffle, bagdasaryan2020backdoor, chen2020backdoor, sun2021fl, Sun2019CanYR, wu2022federated, Wu2020MitigatingBA}, although some have been observed in NLP tasks as well~\citep{bagdasaryan2020backdoor, rieger2022deepsight, nguyen2022flame}.} It is crucial for researchers and practitioners to remain vigilant in exploring the potential of backdoor attacks in various domains and to develop effective defense mechanisms to mitigate their impact.
}

\subsection{Confrontation between Backdoor Attacks and Defenses}
\begin{table*}[hbt!]
\centering
\caption{Backdoor Attacks Strategies and Defense Methodologies in FL}
\label{tab:stat-attack-defense}
\resizebox{0.8\textwidth}{!}{
\begin{tabular}{@{}llll@{}}
\specialrule{0.4pt}{1pt}{1pt}
\multicolumn{3}{c}{Attack Strategies}                                                                                                         & Applicable Defenses                                                         \\ \midrule
\multicolumn{1}{l}{\multirow{6}{*}{Data poisoning}}  & \multicolumn{1}{l}{\multirow{3}{*}{Semantic backdoor}}   & Out-of-distribution        & DeepSight~\cite{rieger2022deepsight} , FLAME~\cite{nguyen2022flame}, Krum~\cite{blanchard2017machine}                                                      \\
\multicolumn{1}{l}{}                                 & \multicolumn{1}{l}{}                                     & In-distribution             & \makecell[l]{FoolsGold~\cite{fung2018mitigating}, VAE~\cite{li2020learning}, AUROR~\cite{shen2016auror}, \\Clustered FL~\cite{sattler2020byzantine} , PCA~\cite{tolpegin2020data}, BaFFle~\cite{andreina2021baffle}, FL-WBC~\cite{sun2021fl} } \\
\multicolumn{1}{l}{}                                 & \multicolumn{1}{l}{}                                     & Noise-instance             & N/A                                                                         \\ \cmidrule(l){2-4} 
\multicolumn{1}{l}{}                                 & \multicolumn{1}{l}{\multirow{3}{*}{Artificial backdoor}} & Single-trigger             & DP \cite{bagdasaryan2020backdoor}, RLR \cite{ozdayi2021defending} , Matching Network~\cite{chen2020backdoor}, Pruning Neurons~\cite{Wu2020MitigatingBA}                                 \\
\multicolumn{1}{l}{}                                 & \multicolumn{1}{l}{}                                     & Distributed-trigger        & \makecell[l]{CRFL~\cite{xie2021crfl}, FLAME~\cite{nguyen2022flame}, RLR~\cite{ozdayi2021defending}, DeepSight~\cite{rieger2022deepsight}, \\Prunning Neurons~\cite{Wu2020MitigatingBA}, FLDetector~\cite{zhang2022fldetector}}                                                  \\
\multicolumn{1}{l}{}                                 & \multicolumn{1}{l}{}                                     & Coordinated-trigger        & N/A                                                                         \\ \midrule
\multicolumn{1}{l}{\multirow{4}{*}{Model poisoning}} & \multicolumn{1}{l}{\multirow{2}{*}{Fully poisoning}}     & Constrain based            & FLARE~\cite{wang2022flare}                                                                       \\
\multicolumn{1}{l}{}                                 & \multicolumn{1}{l}{}                                     & Gradient-replacement based & CAE~\cite{zou2022defending}                                                                         \\ \cmidrule(l){2-4} 
\multicolumn{1}{l}{}                                 & \multicolumn{2}{l}{Partially poisoning}                                               & N/A                                                                         \\ 
\specialrule{0.4pt}{1pt}{1pt}
\end{tabular}
}
\end{table*}



Adversaries and defenders are engaged in a never-ending battle. The conflict between them deepens our understanding of backdoor attacks. Attackers are always looking for ways to make poisoned attacks more covert, effective, and resistant to countermeasures. As shown in Table~\ref{tab:stat-attack-defense}, most defense strategies focus on the scenarios of in-distribution backdoor attacks, in which the adversary simply changes the label of targeted inputs into his expected one. These poisoned samples can appear in the other benign participants' training data. Although defense is often designed against multiple attacks, many attack strategies have not been addressed such as noise-instance, coordinated-trigger, and partially poisoning backdoor attacks.


On the other hand, each countermeasure approach is often applied to a group of attack strategies. Particularly, pre-aggregation methods \tuan{(e.g., Krum~\cite{blanchard2017machine}, FoolsGold~\cite{fung2018mitigating}, and AUROR~\cite{shen2016auror})} seem to be efficient under semantic backdoor attacks. Furthermore, in--aggregation methods are primarily utilized under artificial backdoor attacks, specifically single-trigger attacks. However, the more sophisticated attack strategies such as distributed triggers and coordinated triggers, have not been evaluated under the presence of these defenses.



\section{Challenges and Future Research Directions}
\label{sec:backdoor_future_directions}
\begin{figure}[tbh!]
	\centering
	\includegraphics[width=0.95\textwidth]{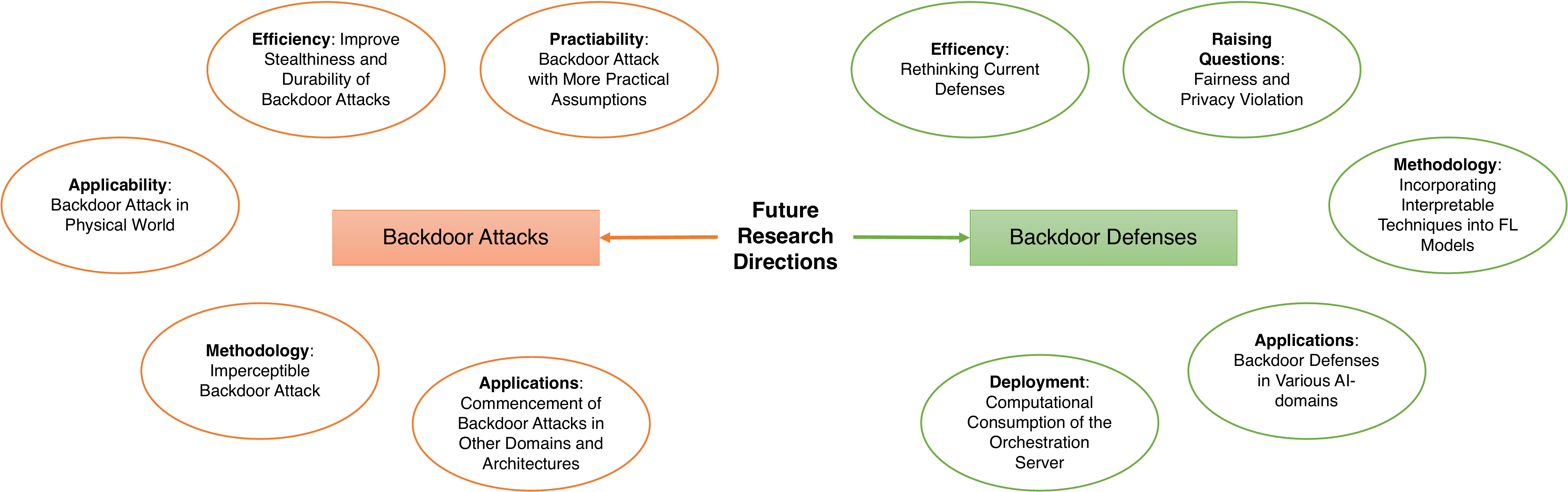}
\caption{Summary of Future Research Directions.}
	\label{fig:future-direction}
\end{figure}

In this section, we first pinpoint aspects for designing a more efficient and robust backdoor attack. Then, we discuss existing disadvantages and corresponding potential research directions for developing backdoor defenses from multi-perspectives. The summary of future research directions is presented in Figure~\ref{fig:future-direction}.
\subsection{Future Research Directions: Backdoor Attacks}

\textbf{Backdoor Attacks with More Practical Assumptions. }Most of the existing backdoor attacks in FL rely on different assumptions, including assumptions about the percentages of compromised clients, the total number of FL clients, and the global distribution of training data. For instance, state-of-the-art attacks~\cite{wang2020attack, HTBD2018} use benign samples drawn from global distribution to manipulate the poisoning dataset. Other attacks~\cite{wang2020attack, fung2018mitigating, PoisonGAN} require continuous participation of compromised clients or a large ratio of malicious clients. This assumption is challenged by~\cite{Shejwalkar2021BackTT} and has shown to be unpractical. Therefore, it would be interesting to explore the possibility of designing attack strategies that require limited assumptions and can be applied in various scenarios, such as in large-scale FL systems with limited knowledge about system operations. To fulfill this purpose, the adversary can exploit leakage information via shared global model~\cite{PoisonGAN, Liu2022TechnicalRA, Xu2020InformationLB} to mimic an auxiliary training dataset align with global data distribution to strengthen the backdoor impact of limited-capability adversaries. Besides, when an adversary controls only a small fraction of participants (i.e., less than $0.1\%$), it can consider designing a single-shot attack and prolong the backdoor durability. 

\textbf{Stealthiness and Durability of Backdoor Attacks. }Most current attacks have not considered stealth or enhanced the stealth by constraining poisoned model updates submitted to the aggregation server. Still, they have not taken the ocular stealth of the attack into account. In the early studies, the trigger is apparent, resulting in poor visual quality, and it can be easily removed by humans~\cite{Doan2021BackdoorAW}. The stealth of the backdoor attacks in FL can be improved from two perspectives. Instead of inserting a small pattern into the original inputs, the trigger should be imperceptible to avoid inspection during the inference procedure. To do this, a learn-able trigger generated by optimizing objective functions~\cite{Gong2022CoordinatedBA, Shafahi2018PoisonFT} or transformation models~\cite{PoisonGAN, Nguyen2021WaNetI} is visually indistinguishable from benign samples. From the model poisoning perspective, the naive scaling-based methods are not stealthy and robust against existing defenses~\cite{Nguyen2022-FLAME, xie2021crfl}. This issue can be addressed by partially poisoning attacks that leverage redundant space within a neural network architecture to covertly implant a backdoor, while still allowing the attacker to scale up the poisoned updates~\citep{Neurotoxin, Zhou2021DeepMP}. In addition, the durability of the backdoor should be intensely considered to avoid the backdoor dilution phenomenon. A robust backdoor attack strategy should well balance stealthiness and durability. 

\textbf{Backdoor Attacks in Physical World. }Current attack strategies typically use an artificial procedure to insert a trigger for a backdoor, such as a small pattern in images during training and testing. However, these attacks can be affected by the loss of the trigger, such as when a camera captures an image from a display or printed photo. The effectiveness of such attacks depends on the location and appearance of the trigger, as discussed in~\cite{Wenger2021BackdoorAA}. Therefore, it is important to evaluate current backdoor threats in physical FL systems. A hybrid attack that works with both digital and physical triggers may be a promising approach for implementing effective backdoor attacks in FL. One of the feasible methods is generating a backdoor dataset with the physical object as a trigger and applying physical transformations to enhance the robustness of the injected backdoor in real-world scenarios~\cite{Wenger2021BackdoorAA, Xue2021RobustBA}.

\textbf{Imperceptible Backdoor Attacks. } The practice of inserting hidden information into images in a way that is imperceptible to the human eye for FL is known as steganography~\citep{hayes2017generating, baluja2017hiding, doan2021lira, jing2021hinet}. This involves concealing a message, image, or file within another message, image, or file without affecting its visible appearance. In the context of FL, steganography could be used to insert data or metadata into images for training ML models while preserving the privacy of sensitive data or transferring it between organizations without revealing its content. Potential approaches to image steganography include applying transformations that preserve visual appearance while encoding additional information, generating adversarial examples with hidden data using machine learning, and developing algorithms for detecting and decoding hidden information in images. The limits of what can be encoded in images while maintaining their visual quality should also be investigated.

\textbf{Commencement of Backdoor Attacks in Other Domains and Architectures. }Backdoor attacks in FL have been mostly studied for image classification~\citep{xie2020dba, Sun2019CanYR, Gong2022CoordinatedBA, ALittleIsEnough2019} and next word prediction~\citep{HTBD2018} tasks. However, existing schemes may not be directly transferable to other domains due to differences in sample nature. Customized strategies may be needed to conduct backdoor attacks in specialized domains such as smart cities~\citep{Zheng2021FederatedLI} or IoT intrusion systems~\citep{Nguyen2020PoisoningIoT}. Some applications of FL, such as environmental monitoring~\citep{Hu2018FederatedRA} and reducing network congestion~\citep{Wang2019InEdgeAI}, lack study on backdoor attacks and require further investigation. HFL is the most attractive land for implanting backdoor attacks since local datasets have the same feature space yet are different from each other and the adversary can easily manipulate the labels for his own training samples. Since VFL and FTL have experienced great development in the industry~\citep{Jing2019QuantifyingTP,Chen2020CommunicationEfficientFD,Wei2022VerticalFL}, 
the presence of a backdoor attack in these scenarios will cause significant concern.
\subsection{Potential Research Directions on Defenses}

\textbf{Differential Privacy in FL.} DP is a framework that protects the privacy of individuals in a dataset by adding noise to the data before it is released or used for analysis. It has been proposed for use in FL~\cite{bagdasaryan2020backdoor, Sun2019CanYR} but has several limitations. DP requires a large number of clients to be effective, as the noise level needs to be high enough to mask the presence or absence of any individual client's data. It may also degrade model performance and may not prevent all types of privacy attacks, such as attribute inference and model inversion. Additionally, DP may not be suitable for all FL scenarios depending on the data being used and the client's privacy requirements. It is important to consider these limitations and trade-offs when using DP in FL settings.

\textbf{Rethinking Current Defenses in FL: Limitations and Uncertainties. }The current defenses in FL have limitations and uncertainties that must be addressed. Firstly, secure aggregation techniques~\cite{bonawitz2017practical}, such as homomorphic encryption and secret sharing, are used in FL to combine model updates from multiple clients while preserving privacy. However, secure aggregation can also make FL systems vulnerable to poisoning attacks as individual updates cannot be inspected. Secondly, the effectiveness of adversarial training in non-IID settings remains uncertain, requiring further research. Finally, the field of FL, including VFL and FTL, is still in its early stages and requires further investigation to fully understand potential backdoor attacks and how to effectively defend against them. To mitigate these concerns, it is important to employ multiple layers of defense mechanisms and continuously monitor and audit the FL process to detect any malicious activity.


\textbf{Backdoor Defenses in Various AI-domains.} Backdoor attacks are generally easier to detect and defend against in the CV domain than in the NLP domain, according to empirical studies. For example, Wan et al.\cite{wan2021robust} found that ASR using the FedAvg algorithm was less than 75\% effective with most defenses when one of ten clients was malicious in CV tasks. However, Yoo et al.\cite{yoo2022backdoor} found that ASR was easily more than 95\% effective on most attacks with most defenses when one of ten clients was malicious in NLP tasks. One reason for this difference may be that detecting NLP backdoors is more difficult. There is increasing interest in using FL in automatic speech recognition~\cite{cui2021federated, dimitriadis2020federated, mdhaffar2021study, guliani2021training}, but the risk of backdoor attacks is a concern that needs to be addressed. Future research may focus on developing effective strategies for defending against and detecting backdoor attacks in the automatic speech recognition domain.

\textbf{Fairness and Privacy Violation. } It is important that the application of a defense mechanism in an FL setting does not impact the fairness among the participating clients. For example, efforts to improve the robustness of FL systems may result in the unfair treatment of honest clients, as their updates may be rejected from the aggregation process if they lie far from the distribution of other updates, as discussed in~\cite{wang2020attack}. This fact raises the question of the compensation between the fairness and robustness of FL systems in the presence of backdoor defenses. Additionally, some defense mechanisms rely on inspecting model updates to study the training data, which can increase the risk of membership inference and model inversion attacks~\citep{rieger2022deepsight, nguyen2022flame}. Therefore, it is important to carefully consider whether a specific defense mechanism is appropriate and to explore more secure defense strategies.

\textbf{Incorporating Interpretable Techniques into FL Models. } Interpretable techniques have been widely studied in the context of single-party ML models, such as decision trees, random forests, gradient-boosted trees, and deep neural networks~\cite{molnar2020interpretable, rudin2019stop, montavon2018methods, samek2019towards}. Most of these techniques have been developed to provide transparency into the decision-making processes of these models, with the goal of enhancing their interpretability and usability. However, their application to FL is relatively new. By providing transparency into the decision-making processes of FL models, interpretability techniques can help detect malicious clients and prevent backdoor attacks. For instance, studies show that saliency maps can reveal hidden triggers in single-party models and demonstrate the effectiveness of different defense methods against backdoors~\citep{fang2022backdoor}. Similarly, visualization techniques can help to identify regions of the model's input space that are particularly susceptible to backdoor attacks and provide a way to test and validate the robustness of FL models.

\textbf{Computational Consumption of the Orchestration Server. }The deployment of defense mechanisms in FL requires significant computational resources, and it's crucial to ensure that it doesn't exceed the capacity of the orchestration server. Existing defense mechanisms often overlook the limitation of computational resources, leading to time delays and energy consumption. In future research, it's important to minimize resource consumption while deploying defense mechanisms in FL. For instance, for FL systems with a small number of clients, the local models can be verified one by one, but when the number of clients increases, this approach becomes impractical and consumes vast amounts of time and energy. An alternative solution is to deploy FL with multiple servers, distributing the task of verifying updates among them, which reduces resource consumption but brings new challenges such as communication costs and privacy leakage. Another promising solution is combining FL with blockchain technology, as proposed in~\cite{chen2021robust}, where clients upload updates to verifiers who select benign updates by voting and then aggregate and write the selected updates to blocks through the blockchain network.


\subsection{Discussion}
\textbf{Exploring the Practical Benefits of Backdoor Attacks in Federated Unlearning.} We often consider backdoor attacks as a great threat in FL while ignoring its potential advantages. Indeed, a backdoor attack demonstrated its sake under the unlearning scenario, which is a technique in FL~\cite{wu2022federated, liu2021federaser, wang2022federated} focusing on removing or revoking access to data, participants, or parts of the model, with the goal of improving the integrity and accuracy of the model. Backdoor triggers are utilized as an evaluation tool to assess the effectiveness of unlearning methods~\cite{halimi2022federated}. The client, who wants to opt out of the federation, uses a dataset that contains a fraction of samples with inserted backdoor triggers, making the global FL model vulnerable to the backdoor trigger. The goal of the unlearning process is to produce a model that decreases accuracy on samples with backdoor triggers while preserving good performance on clean samples. Future research on unlearning needs to focus more on investigating the impact of backdoor attack methods on model privacy and security.

\textbf{Investigating Various Backdoor Injection Strategies in Multi-Group FL. }Existing works often consider the homogeneous backdoor attack, in which the malicious participants have a common attack objective~\cite{wang2020attack, xie2020dba, HTBD2018, rieger2022deepsight}. This approach is not always relevant in the physical world. This raises a great concern about backdoor effects caused by multiple adversaries with different backdoor targets. 
For example, consider a model that recognizes two-digit numbers. It is possible to inject two new backdoor tasks into the model: one that sums up the digits and another one that multiplies them. Then, various endeavors from distinct backdoor tasks can be complementary or detrimental to one another. Moreover, the appearance of multiple backdoor tasks may have varying effects on the performance of the FL model, and it is important for future research to uncover these effects in order to improve the security and privacy of FL models. This research can aid in the development of better methods for detecting and mitigating backdoor attacks in FL, thus improving the overall integrity and robustness of the model.


\textbf{Integrating Multiple Defense Mechanisms in FL. } \tuan{Previous works~\cite{Nguyen2022-FLAME, rieger2022deepsight}} used a combination of methods in the Pre-AD and In-AD phases to mitigate backdoor attacks. These methods involve two layers: the first layer detects and excludes models that contain a well-trained backdoor, while the second layer uses a different approach in the In-AD phase to mitigate the attack. In future research, a combination of methods from different defense phases, such as Pre-AD and Post-AD, In-AD and Post-AD, or Pre-AD, In-AD, and Post-AD, can be studied to further improve the defense against backdoor attacks in FL.

\section{Conclusion}
\label{sec:conclusion}

In summary, backdoor attacks in FL pose a significant threat to the security and privacy of FL systems. These attacks can be triggered in various ways, including artificial and semantic triggers, and can be launched by a single client or a group of clients. To defend against these attacks, various approaches have been proposed, including pre-aggregation defenses, in-aggregation defenses, and post-aggregation defenses. Each of these approaches has its own advantages and limitations, and their effectiveness depends on the specific characteristics of the attack. Moreover, the robustness of these defenses in the face of various types of attacks, particularly in non-IID scenarios, remains an open research question. In the future, it will be important to continue developing more robust defense techniques that are effective against semantic backdoor attacks, improving the efficiency of defense techniques, studying the effectiveness of defenses under realistic attack scenarios, examining the impact of data heterogeneity on backdoor attacks and defenses, and investigating the impact of system-level factors on backdoor attacks and defenses. By addressing these research areas, it will be possible to make progress in understanding and addressing the risks of backdoor attacks in federated learning systems and to develop more secure and effective defense strategies against a wide range of attacks. It is also important to consider the potential for physical backdoor attacks and to explore potential defenses against these types of attacks. In addition, research on the effectiveness of backdoor defenses in specific AI domains, such as automatic speech recognition, could be valuable in developing targeted and effective protection mechanisms.

\section*{CrediT Authorship Contribution Statement}
\textbf{Thuy Dung Nguyen:} Methodology, Visualization, Writing - Original Draft, Writing - Review \& Editing, \textbf{Minh Tuan Nguyen:} Methodology, Visualization, Writing - Original Draft, Writing - Review \& Editing, \textbf{Phi Le Nguyen:} Writing - Review \& Editing, \textbf{Huy Hieu Pham:} Writing - Review \& Editing, \textbf{Khoa Doan:} Writing - Review \& Editing, \textbf{Kok-Seng Wong:} Conceptualization, Project administration, Supervision, Writing - Review \& Editing.

\section*{Declaration of Competing Interest}
The authors declare that they have no known competing financial interests or personal relationships that could have appeared to influence the work reported in this paper.

\section*{Acknowledgment}
This work was supported by VinUni-Illinois Smart Health Center (VISHC), VinUniversity.





\bibliographystyle{elsarticle-num}
\bibliography{main}







\end{document}